%% file: example_paper.tex
\documentclass{article}

\usepackage{microtype}
\usepackage{graphicx}
\usepackage{subfigure}
\usepackage{booktabs} 

\usepackage{hyperref}

\usepackage[accepted]{icml2023}

\usepackage{amsmath}
\usepackage{amssymb}
\usepackage{mathtools}
\usepackage{amsthm}

\usepackage[capitalize,noabbrev]{cleveref}

\theoremstyle{plain}
\newtheorem{theorem}{Theorem}[section]
\newtheorem{proposition}[theorem]{Proposition}

\theoremstyle{definition}
\newtheorem{definition}[theorem]{Definition}
\newtheorem{assumption}[theorem]{Assumption}

\theoremstyle{remark}
\newtheorem{remark}[theorem]{Remark}

\usepackage[textsize=tiny]{todonotes}

\usepackage{makecell}
\usepackage{bbm}
\usepackage{booktabs}  
\usepackage{multirow}
\usepackage{diagbox}

\usepackage[textsize=tiny]{todonotes}

\icmltitlerunning{ConCerNet: A Contrastive Learning Based Framework}
\begin{document}

\twocolumn[
\icmltitle{ConCerNet: A Contrastive Learning Based Framework for Automated Conservation Law Discovery and Trustworthy Dynamical System Prediction}





\begin{icmlauthorlist}
\icmlauthor{Wang Zhang}{mit}
\icmlauthor{Tsui-Wei Weng}{ucsd}
\icmlauthor{Subhro Das}{mitibm}
\icmlauthor{Alexandre Megretski}{mit}
\icmlauthor{Luca Daniel}{mit}
\icmlauthor{Lam M. Nguyen}{ibmresearch,mitibm}
\end{icmlauthorlist}

\icmlaffiliation{mit}{Massachusetts Institute of Technology, Cambridge, MA, USA}
\icmlaffiliation{ucsd}{University of California, San Diego, CA, USA}
\icmlaffiliation{mitibm}{MIT-IBM Watson AI Lab, IBM Research, Cambridge, MA, USA}
\icmlaffiliation{ibmresearch}{IBM Research, Thomas J. Watson Research Center, Yorktown Heights, NY, USA}

\icmlcorrespondingauthor{Wang Zhang}{wzhang16@mit.edu}
\icmlcorrespondingauthor{Lam M. Nguyen}{LamNguyen.MLTD@ibm.com}

\icmlkeywords{Machine Learning, ICML}

\vskip 0.3in
]



\printAffiliationsAndNotice{}  


\begin{abstract}
Deep neural networks (DNN) have shown great capacity of modeling a dynamical system; nevertheless, they usually do not obey physics constraints such as conservation laws. This paper proposes a new learning framework named \textbf{ConCerNet} to improve the trustworthiness of the DNN based dynamics modeling to endow the invariant properties. \textbf{ConCerNet} consists of two steps: (i) a contrastive learning method to automatically capture the system invariants (i.e. conservation properties) along the trajectory observations; (ii) a neural projection layer to guarantee that the learned dynamics models preserve the learned invariants. We theoretically prove the functional relationship between the learned latent representation and the unknown system invariant function. Experiments show that our method consistently outperforms the baseline neural networks in both coordinate error and conservation metrics by a large margin. With neural network based parameterization and no dependence on prior knowledge, our method can be extended to complex and large-scale dynamics by leveraging an autoencoder. 
\end{abstract}

\input{1_intro.tex}
\input{2_background.tex}

\input{3_method.tex}
\input{4_theory.tex}
\input{5_experiment.tex}
\input{6_discussion.tex}

\newpage
\bibliography{example_paper}
\bibliographystyle{icml2023}

\newpage
\appendix
\onecolumn
\input{7_appendix.tex}


\end{document}

%% file: 1_intro.tex
\vspace{-0.2in}
\section{Introduction}


Many critical discoveries in the world of physics were driven by distilling the invariants from observations. For instance, the Kepler laws were found by analyzing and fitting parameters for the astronomical observations, and the mass conservation law was first carried out by a series of experiments. However, such discoveries usually require extensive human insights and customized strategies for specific problems. This naturally raises a question:
\vspace{-0.05in}
\begin{quote}
\textbf{Q1:}
\textit{Can we learn conservation laws from real-world data in an automated fashion?}  
\end{quote}
\vspace{-0.05in}
%
Recent works on automatic discovery of scientific laws try to answer the above question by proposing symbolic regression \cite{udrescu2020aiFeynman}. The idea is to recursively fit the data to different combinations of pre-defined function operators. However, these methods' implicit dependence on human knowledge (i.e. function class and complexity) and high computational cost limit their application to small physics equations,
making them less scalable and incapable of handling general systems with complicated dynamics.

On the other hand, the approach of data-driven dynamical modeling tries to learn a dynamical system from data, which often generate models that are prone to violation of physics laws as demonstrated in \citep{greydanus2019hamiltonian}. This motivates a second question:
\vspace{-0.05in}
\begin{quote}
\textbf{Q2:}
\textit{Can we learn a dynamical system that is trustworthy (i.e. obey physics laws)?}  
\end{quote}
\vspace{-0.05in}
There has been a recent line of work trying to answer this question by actively constructing dynamics models that obey physical constraints. For example, \citet{greydanus2019hamiltonian} enforces the Hamiltonian to be conserved in Hamiltonian systems, \citet{cranmer2020lagrangian} further extends it to Lagrangian dynamics. As these works focus on specific systems, there is yet a method to be designed for preserving the conservation quantities in \textit{general} dynamical systems.

\begin{figure*}[htb!]
\centering
\includegraphics[width=1.4\columnwidth]{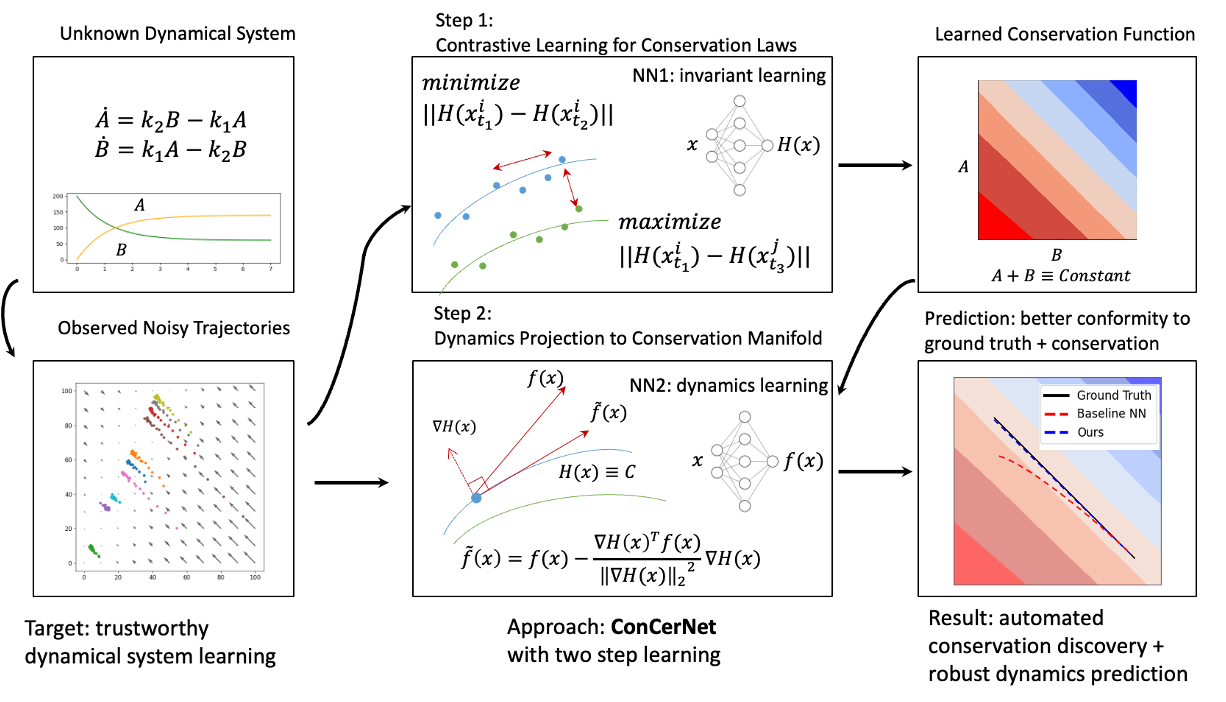}
\caption{\textbf{ConCerNet} pipeline to learn the dynamical system conservation and enforce it in simulation. A contrastive learning framework is proposed to extract the invariants across trajectory observations, then the dynamical model is projected to the invariant manifold to ensure the conservation property.}
\label{fig:pipeline}
\end{figure*}

In this work, we aim at answering the two questions \textbf{Q1} and \textbf{Q2} jointly by introducing a new framework \textbf{ConCerNet}\footnote{Source code available at \url{https://github.com/wz16/concernet}.}, which is a novel pipeline to learn trustworthy dynamical systems $f(x)$ based on the automatic discovery of the system's invariants $H(x)$ from data. \textbf{ConCerNet} consists of two steps: (i) contrastively learning $H(x)$ as low-dimensional representation as the invariant quantity for the system; (ii) learning $f(x)$ to approximate the system dynamics with a correction step. This ensures the learned dynamics to automatically preserve the learned $H(x)$. An overview of ConCerNet method pipeline is shown in \Cref{fig:pipeline}.

We summarize our main contributions as follows:
\begin{enumerate}
    \item We provide a novel contrastive learning perspective of dynamical system trajectory data to capture their invariants. We design a \textbf{Square Ratio Loss} function as the contrastive learning metric. 
    To the best of the authors' knowledge, this is the first work that studies the discovery of conservation laws for general dynamical systems through contrastive learning.


    \item We propose a \textbf{Projection Layer} to impose conservation of the invariant function for dynamical system trajectory prediction, preserving the conservation quantity during dynamics modeling.
    
    \item Leveraging the above components, we establish a generic learning framework for dynamical system modeling named \textbf{ConCerNet} (CONtrastive ConsERved Network). It provides robustness in prediction outcomes and flexibility for application in a wide range of dynamical systems that mandate conservation properties. 

    \item Under mild conditions, we prove the local minimum property of \textbf{Square Ratio Loss}, theoretically bridging the relationship between the learned latent representation and original conservation function.
    
    \item We conduct extensive experiments to demonstrate the efficacy of \textbf{ConCerNet}. Our contrastive learning method automatically discovers physical conservation laws, and the coordinate error/conservation violation of \textbf{ConCerNet} are much smaller than the baseline neural networks in dynamics prediction.
    
    
\end{enumerate}


%% file: 2_background.tex
\section{Background and Related Work}

\subsection{Contrastive Learning}
Unlike discriminative models that explicitly learn the data mappings, contrastive learning aims to extract the data representation implicitly by comparing among examples. The early idea dates back to the 1990s \citep{bromley1993signature} and has been widely adopted in many areas. One related field to our work is metric learning \citep{chopra2005metric,harwood2017smart,sohn2016improvedmetric}, where the goal is to learn a distance function or latent space to cluster similar examples and separate the dis-similar ones. 

Contrastive learning has been a popular choice for self-supervised learning (SSL) tasks recently, as it demonstrated its performance in many applications such as computer vision \citep{chen2020simple,he2020momentum,ho2020adversarialcontrastive,tian2020contrastive} and natural language processing \citep{wu2020clear, gao-etal-2021-simcse}. There are many existing works related to the design of contrastive loss \citep{oord2018cpc,chen2020simple,chen2020big}. 
Prior to our work to use contrastive learning in exploiting temporal correlation, \citet{hyvarinen2016unsupervised,hyvarinen2017nonlinear} propose time contrastive learning to discriminate time segments and \citet{eysenbach2022contrastive} estimates Q-function in reinforcement learning by comparing state-action pairs.

\subsection{Deep Learning based Dynamical System Modeling}
Constructing dynamical system models from observed data is a long-standing research problem with numerous applications such as forecasting, inference and control. System identification (SYSID) \citep{Ljung1974,ljung1998systemid,keesman2011system} was introduced 5-6 decades ago and designed to fit the system input-output behavior with choice of lightweight basis functions. In recent years, neural networks became increasingly popular in dynamical system modeling due to its representation power. In this paper, we consider the following neural network based learning task to model an (autonomous and continuous time) dynamical system:
\begin{equation}\label{eqn:continuous_dynamics}
    f_{\theta}(x(t)) \sim \dot{x}(t)\equiv \frac{dx(t)}{dt}
\end{equation}
where $x\in \mathbb{R}^n$ is the system state and $\dot{x}$ is its time derivative. $f_\theta: \mathbb{R}^n\rightarrow \mathbb{R}^n$ denotes the neural network model $f$ with parameter $\theta$,  and $f_\theta$ is used to approximate dynamics evolution.

The vanilla neural networks learn the physics through data by minimizing the step prediction error, without a purposely designed feature to honor other metrics such as conservation laws. One path to address this issue is to include an additional loss in the training \citep{singh2021contraction, wu2020enforcing,richards2018lyapunov,wang2020turbulent}; however, the soft Lagrangian treatment does not guarantee the model performance during testing. Imposing hard constraints upon the neural network structures is a more desirable approach, where the built-in design naturally respect certain property regardless of input data. Existing work includes: \citet{kolter2019stabledeep} learns the dynamical system and a Lyapunov function to ensure the exponential stability of predicted system; Hamiltonian neural network (HNN, \citet{greydanus2019hamiltonian}) targets at the Hamiltonian mechanics, directly learns the Hamiltonian and uses the symplectic vector field to approximate the dynamics; Lagrangian neural network (LNN, \citet{cranmer2020lagrangian}) extends the work of HNN to Lagrangian mechanics. Although the above models are able to capture certain conservation laws under specific problem formulations, they are not applicable to general conserved quantities (e.g. mass conservation). This motivates our work in this paper to propose a contrastive learning framework in a more generic form that is compatible of working with arbitrary conservation.

\subsection{Learning with Conserved Properties}
Automated scientific discovery from data without prior knowledge has attracted great interest to both communities in physics and machine learning. Besides the above-mentioned HNN and LNN, a few recent works \citep{zhang2018machine,liu2021PRL, ha2021trajconsv,liu2022aipoincare2, udrescu2020aiFeynman} have explored automated approaches to extract the conservation laws from data. Despite the promising results, the existing methods are mostly based on symbolic regression, and they suffer from limitations including poor data sampling efficiency and rely on prior knowledge and artificial preprocessing. Often they use search algorithms on pre-defined function classes, hereby are difficult to extend to larger and more general systems. We show the method applicability comparison in \Cref{tab:method_overview}. To the best of our knowledge, our work is the first time to study automated conservation law discovery: 1. for general dynamical system with arbitrary conservation property without any prior knowledge 2. through the lens of contrastive learning.



%% file: 3_method.tex
\section{Proposed Methods: ConCerNet}

\subsection{Step 1: Learning Conservation Property from Contrastive Learning}\label{sec:method step 1}
\input{tables/method_compare.tex}

In the practice of dynamical system learning, the dynamics data is usually observed as a set of trajectories of system state $\{x^i_{t}\in \mathbb{R}^n\}_{i=1, t=1}^{N,T}$, where $i$ denotes the trajectory index of total trajectory number $N$ and $t$ is the time step with the number of total time step $T$. Let $\{x^i_{1}\}_{i=1}^N$ be the initial conditions and assume they have different conservation values. Let $H_{\theta_c}: \mathbb{R}^n \rightarrow \mathbb{R}^m$ be the $m$-dimensional latent mapping function parameterized by the neural network. Following the convention of classical contrastive learning, the system states within the same trajectory are considered ``in the same class''  and with the same value of conservation properties. Since the invariants are well conserved along the trajectory and differ among different trajectories, we aim to find the conservation laws that naturally serve as each class's latent representation. 



As a metric to persuade similar latent representation within the same trajectory and encourage discrepancy between different trajectories, we introduce the $\textbf{Square Ratio Loss}$ (SRL) defined in the following as the contrastive loss function:
\begin{align}\label{eqn:ratio loss discretized}
    & \mathcal{L_\text{SR}}  = \frac{1}{NT}\sum_{i=1}^{N} \sum_{t_1=1}^{T} \nonumber  \\
    & \left[ \frac{ \sum_{t_2=1}^{T} \lVert H_{\theta_c}(x_{t_1}^i)-H_{\theta_c}(x_{t_2}^i)\rVert^2}{\sum_{j=1}^{N} \sum_{t_3=1}^{T} \lVert H_{\theta_c}(x_{t_1}^i)-H_{\theta_c}(x_{t_3}^j)\rVert^2}\right].
\end{align}

In each fraction, the denominator summarizes the squared Euclidean norm in the latent space between the anchor point and all other points, the numerator only summarizes between the anchor point and points within the same trajectory. Therefore, we call this loss function design ``square ratio loss''. Intuitively, minimizing this loss function will decrease the latent discrepancy between points within the same trajectory and increase the distance between different trajectories.

To notice, in common contrastive learning settings like SimCLR \citep{chen2020simple}, the latent space is usually measured by Cosine distance between point pairs, rather than the Euclidean distance. Besides, contrastive learning is a classification task while our metric is pure value comparison. Further, we compare one point to a group of points in the same class, this is similar to the NaCl loss in \cite{ko2022revisiting} with many similar points to the anchor point. We choose SRL as metric for a few reasons: 1. In contrast with classification objects with discretized sampling distribution space, the dynamical system lives in a continuous space, Euclidean distance comparison intuitively works better. 2. SRL achieves good experimental performance, as we will show in \Cref{sec:experiments}. 3. Through the latter theoretical analysis in \Cref{thm:formal}, we can prove the optimization property of SRL that draws the relationship between the learned representation and the exact conservation law.

\subsection{Step 2: Enforcing Conservation Invariants in Dynamical Modeling}\label{sec:method step 2}

Once the conservation term $H_{\theta_c}(x)\in\mathbb{R}^m$ is contrastively learned or given from prior knowledge, we attempt to enforce the predicted trajectory along the conservation manifold in the simulation stage, $s.t. \textbf{ } \frac{dH_{\theta_c}(x) }{dt}=0$. In the continuous dynamical system like \Cref{eqn:continuous_dynamics}, we can project the nominal neural network output $f_{\theta_d}(x)$ onto the conservation manifold by eliminating its parallel component to the normal direction of the invariant planes (i.e. $\nabla_x H_{\theta_c}(x)$). Let $G=\nabla_x H_{\theta_c}(x)\in \mathbb{R}^{n\times m}$, we define the projected dynamical model $\tilde{f}_{\theta_d}(x)$ as following:
\vskip -0.2in
\begin{align}
    & \tilde{f}_{\theta_d}(x) \coloneqq \text{Projection}\left(f_{\theta_d}(x), \{ f: G^\top f =0 \}\right) \nonumber \\
    & = f_{\theta_d}(x) - G(G^\top G)^{-1}G^\top f_{\theta_d}(x)\nonumber \\
    & = f_{\theta_d}(x)- \sum_{i=1}^m (G^\perp_i)^\top  f_{\theta_d}(x) G^\perp_i,
\end{align}

where the second equality is the standard orthogonal projection equation and the third equality indicates we use the Gram-Schmidt process to solve it in practice. Here $G^\perp$ denotes the orthonormalized matrix from $G$ calculated by Gram–Schmidt process and $ G^\perp_i $ is the $ith$ component. The projected dynamics function naturally satisfies $ (\nabla_x H_{\theta_c}(x))^\top f = 0 $ and therefore guarantees $H_{\theta_c}(x)$ to be constant during prediction. The intuitive diagram of projection with one conservation term (i.e. $m=1$) is shown in \Cref{fig:pipeline}.

For dynamical system learning, we assume the system time derivative is observable (with noise) for simplicity. The loss function is the mean square loss between the neural network prediction and the system time derivative, as shown in \Cref{eqn:dynamicsloss}. For real-world problems with only discretized state observations, we can approximate the derivative by time difference or bridge the continuous system and discretized data by leveraging neuralODE \citep{chen2018neuralode}. 
\vspace{-0.04in}
\begin{equation}\label{eqn:dynamicsloss}
    \mathcal{L_\text{dyn}}  = \mathbb{E}_{ \substack{x}} \left[ \lVert \tilde{f}_{\theta_d}(x)- \dot{x} \rVert^2\right].
\end{equation}                                         

%% file: tables/method_compare.tex
\begin{table}[t]
\caption{Applicability comparison of automatic methods to recover conservation laws.}
\label{tab:method_overview}
\vskip -0.2in
\begin{center}
\scalebox{0.7}{
\begin{tabular}{lcccr}
\toprule
Method & Applicability & \makecell{Do not need pre-  processing \\$/$prior knowledge} \\
\midrule
\makecell{Symbolic Reg.} & Simple conservation & $\times$\\
HNN/LNN & Hamiltonians/Lagrangians& $\surd$\\
ConCerNet   & General conservation & $\surd$\\
\bottomrule
\end{tabular}
}
\end{center}
\vskip -0.2in
\end{table}

%% file: 4_theory.tex
\section{Theory}\label{sec:theory}

In this section, we rigorously formulate the contrastive learning problem from \Cref{sec:method step 1} and study the latent function property in the case of the SRL loss.

\begin{definition}\label{def:inverse neighborhood}
    For function $g(\cdot):\mathcal{X} \rightarrow \mathbb{R}^1$ defined on a compact and convex set $\mathcal{X}\subset\mathbb{R}^n$, $y\in \mathbb{R}^1$, $\epsilon>0$, define the \textbf{preimage} $g^{-1}_{\mathcal{X}}(y,\epsilon)=\{ x | x\in \mathcal{X}, ||g(x)-y||\leq \epsilon \}$. The preimage represents the set of elements in $\mathcal{X}$ mapped to the $\epsilon$ ball centered at $y$ in the image space. $\epsilon$ is called image neighborhood diameter. 
\end{definition}



\begin{definition}\label{def:csrl}
    For a given implicit $g(\cdot)$ and state set $\mathcal{X}$, define the \textbf{Integral Square Ratio Loss (ISRL)} as a function of neighborhood diameter $\epsilon$ and target function $h(\cdot):\mathbb{R}^n \rightarrow \mathbb{R}^1$:
    \begin{align}\label{eqn:continuous SRL}
    \mathcal{L}_{ISR}(h(\cdot),\epsilon) & = \int\displaylimits_{\mathcal{X}}
    \frac{
    \int_{g^{-1}_{\mathcal{X}}(g(x),\epsilon)} (h(x)-h(x'))^2 dx'
    }{
    \int\displaylimits_{\mathcal{X}} (h(x)-h(x''))^2 dx''
    } dx.
\end{align}
\end{definition}

\begin{remark}\label{remark:H dim}
To facilitate the notation and problem formulation, we convert the discretized SRL from \Cref{eqn:ratio loss discretized} into the continuous and surrogate version of SRL in \Cref{eqn:continuous SRL}. To notice, they are not equivalent: 1. In the discretized version, the numerator summation set is the points within the same trajectory, it is a ``partial preimage'' being a subset of the preimage integral area in the continuous loss. In many cases, we only have partial preimage due to observation interval or trajectory length. For the rest of the analysis, we consider the ideal case with access to full preimage. However, we can perform a similar analysis on the partial preimage case and the final functional relationship will be confined to the union of the observed partial preimage. 2. In practice, there exists noise in observation. The continuous version explicitly counts in the factor and summarizes over all the preimage within $\epsilon$ diameter to the anchor point. 3. The actual system might have more than one conservation law and \textbf{ConCerNet} allows more than one dimension for $H_{\theta_c}(\cdot)$. In this section, we only consider $g(\cdot)$ and $h(\cdot)$ being one-dimensional. For a system with multiple conservation laws, we can pick one conservation term. In fact, we cannot even guarantee to find the exact conservation function for such systems, as any combination of different conservation functions is conserved as well.

\end{remark}
\newpage
\begin{definition}\label{def:injective}
    Let $\mathcal{X}\subset \mathbb{R}^n$ be a compact and convex set with positive volume (w.r.t. the Lebesgue measure). Let $\mathcal{F}$ denote the space of all square integrable\footnote{The element in $L^2$ space is an equivalent class of functions equals to each other almost everywhere, here we use one function $f$ to represent the equivalent class, as all equivalent function have the same integral result on positive volume.} real-valued functions $f:\mathcal{X}\rightarrow \mathbb{R}$. \\
    For function $f_1(\cdot),f_2(\cdot) \in\mathcal{F}$, we say that $f_2(\cdot)$ is \textbf{``functional injective''} with respect to $f_1(\cdot)$ when $\forall x_1,x_2 \in \mathcal{X},  f_1(x_1)=f_1(x_2) \implies f_2(x_1)=f_2(x_2)$. \\
    Let $\mathcal{F}_{inj}(f_1)$ 
     denote ``functional injective set'' with respect to $f_1(\cdot)$, the set of all the functions in $\mathcal{F}$ which are functional injective to $f_1(\cdot)$.
     \begin{remark}
    $f_2(\cdot)$ being ``functional injective'' to $f_1(\cdot)$ is equivalent to saying that there exists another function $f_3(\cdot)$, s.t. $f_2(\cdot)$ can be written as a composition of $f_3(\cdot)$ and $f_1(\cdot)$: $f_2(x)=f_3(f_1(x))$.
    \end{remark}
      For functions $f_1(\cdot),f_2(\cdot) \in\mathcal{F}$, we say that $f_2(\cdot)$ is \textbf{``skew-symmetric functional injective''} with respect to $f_1(\cdot)$, we have $\mathbb{E}[f_2|f_1]=0$.\footnote{
      With $\mathcal{X}$ as the sampling space of random variable $x$, we define the probability measure $P$, s.t. $\forall A\subset \mathcal{X}$, $P(A)=|A|/|\mathcal{X}|$, where $|\cdot|$ represents the $n$-dimensional Lebesgue measure. Consider $f_1$ and $f_2$ as two random variables, the conditional expectation $\mathbb{E}[f_2|f_1]$ is a function that takes the value of $\mathbb{E}[f_2|f_1=r]$ whenever $f_1=r$, i.e., $\phi(r)=\int f_2p(f_2|f_1=r) df_2=\int_{f^{-1}_1(r)}f_2(x)p(f_2(x)|f_1(x)=r)dx$. $p(f_2(x)|f_1(x)=r)$ has the Dirac delta form if $P(f^{-1}_1(x)=r)=0$.}\\
      Let $\mathcal{F}_{skewsym\text{-}inj}(f_1)$ 
     denote ``skew-symmetric functional injective set'' with respect to $f_1(\cdot)$, the set of all the functions in $\mathcal{F}$ which are skew-symmetric functional injective to $f_1(\cdot)$. 
 \end{definition}

\begin{proposition}\label{prop:vectorspace}
$\mathcal{F}$ defined in \Cref{def:injective} is a vector space over field $\mathbb{R}$. For square integrable function $f_1(\cdot):\mathcal{X}\rightarrow\mathbb{R}^1$,  $\mathcal{F}_{inj}(f_1)$ and $\mathcal{F}_{skewsym\text{-}inj}(f_1)$ are complemented subspaces of $\mathcal{F}$. 
\end{proposition}

\begin{definition}\label{def:local_min}
    Let $\mathcal{L}(\cdot): \mathcal{F}\rightarrow \mathbb{R}^1$ be a function on $ \mathcal{F}$. Let $\mathcal{F}_1$ and $\mathcal{F}_2$ be complemented subspaces of $\mathcal{F}$. We say $\mathcal{L}$ reaches a ``directional local minimum'' for any perturbation along $\mathcal{F}_2$ at $f_1(\cdot) \in\mathcal{F}_1$, if $ \forall f_2\in\mathcal{F}_2$ not being a zero function, $ \exists\overline{\delta}>0,\text{ such that } \mathcal{L}(f_1(\cdot)+\delta f_2(\cdot))>\mathcal{L}(f_1(\cdot)),\forall \delta < \overline{\delta}$.
\end{definition}

\begin{definition}
    For function $f_1(\cdot)$ and $f_2(\cdot)$ defined on $\mathcal{X}$, we say $f_2(\cdot)$ is ``C-relative Lipschiz continuous'' to  $f_1(\cdot)$ if $\forall x_1,x_2\in \mathcal{X}, ||f_2(x_1)-f_2(x_2)||\leq C ||f_1(x_1)-f_1(x_2)||$ for some constant $C$.
\end{definition}


With the above preparation, we are ready to introduce the main theoretical result of this paper, which studies the local minimum property of the loss function when $h(\cdot)$ is ``functional injective'' to $g(\cdot)$. We make three moderate assumptions in the following. Assumption \ref{ass:rela_continuous} caps the variation magnitude between $h(\cdot)$ and $g(\cdot)$. \Cref{ass:gen_variance} and \Cref{ass:epsilon} require the image neighborhood diameter $\epsilon$ to be small. The proof of \Cref{thm:formal} is delayed to \Cref{appendix:proof}.

\begin{theorem}\label{thm:formal}
    For a compact and convex set $\mathcal{X}\subset \mathbb{R}^n$ of positive volume, consider a non-constant continuously differentiable (thus also square integrable) function $g(\cdot): \mathcal{X} \rightarrow \mathbb{R}^1$ and corresponding preimage operator $g^{-1}_{\mathcal{X}}(\cdot,\epsilon)$ (defined in \Cref{def:inverse neighborhood}) with diameter $\epsilon>0$. Let $\mathcal{F}$ be the functional space of square integrable real-valued functions defined on $\mathcal{X}$. Let $\mathcal{L}_{ISR}$ be the integral square ratio loss from \Cref{def:csrl}. Assume $h(\cdot)\in \mathcal{F}_{inj}(g)$ and $\epsilon>0$ satisfy \Cref{ass:rela_continuous}, \ref{ass:gen_variance} and \ref{ass:epsilon}. Then $\mathcal{L}_{ISR}$ reaches a directional local minimum (defined in \Cref{def:local_min}) at $h(\cdot)$ for any perturbation from $\mathcal{F}_{skewsym\text{-}inj}(g)$ (defined in \Cref{def:injective}) . 
\end{theorem}

\begin{assumption}\label{ass:rela_continuous}
    $h(\cdot)$ is $C_1$-relative Lipschiz continuous to $g(\cdot)$ and $g(\cdot)$ is $C'_1$-relative Lipschiz continuous to $h(\cdot)$.
\end{assumption}


\begin{assumption}\label{ass:gen_variance}
    The image neighborhood diameter $\epsilon$ (defined in \Cref{def:csrl}) is small enough s.t. $\forall x \in \mathcal{X}$,\\
    $\epsilon < \sqrt{\frac{\int_{\mathcal{X}} (h(x)-h(x''))^2 dx''}{\int_{\mathcal{X}} 1 dx''}}/(C_1\sqrt{5C_2})$, where $C_1$ is the Lipschitz continuous constant defined in \Cref{ass:rela_continuous} and $C_2 = \frac{\max_{x_1}\int_{\mathcal{X}} (h(x_1)-h(x''))^2 dx''}{\min_{x_2}\int_{\mathcal{X}} (h(x_2)-h(x''))^2 dx''}$.
\end{assumption}


\begin{assumption}\label{ass:epsilon}
    $\epsilon$ is small enough s.t. for $\forall x \in \mathcal{X}$, 
    \begin{align*}
        C_1\epsilon \frac{\int_{x'\in g^{-1}_{\mathcal{X}}(g(x),\epsilon)}(h(x)-h(x'))dx'}{{\int_{x'\in g^{-1}_{\mathcal{X}}(g(x),\epsilon)}(h(x)-h(x'))^2}dx'}<\int_{x\in\mathcal{X} } 1dx.
    \end{align*}
\end{assumption}

\vskip 0in
\begin{figure}[htb!]
\centering
\includegraphics[width=0.8\columnwidth]{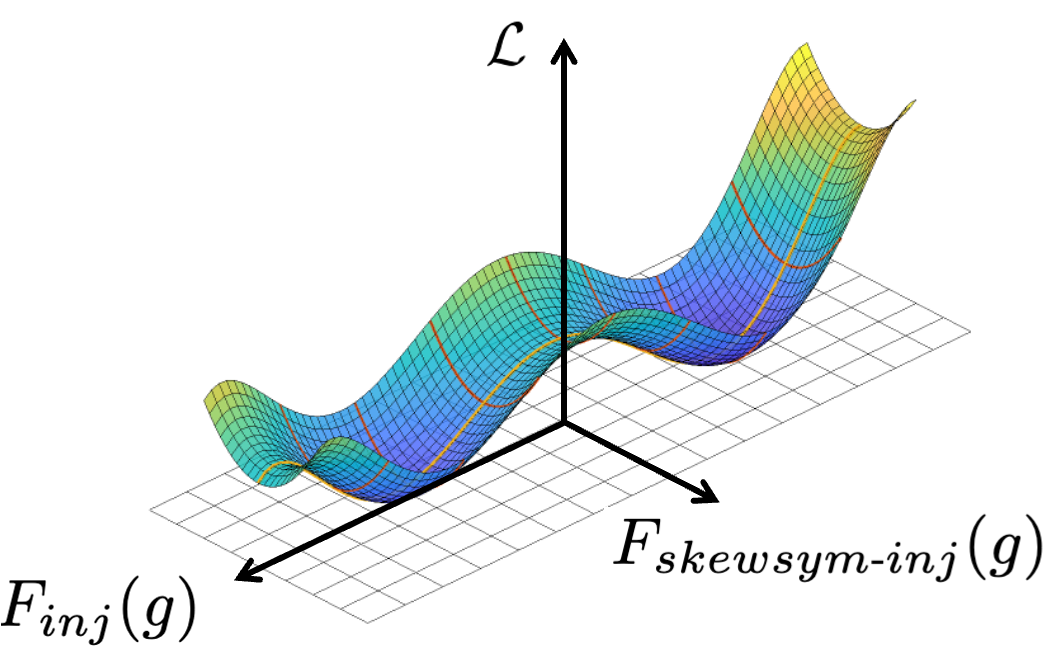}
\caption{Illustrative Diagram for \Cref{thm:formal}. The yellow curve denotes $h(\cdot)\in \mathcal{F}_{inj}(g)$, the red curves denote the perturbations in $\mathcal{F}_{skewsym\text{-}inj}(g)$.}
\label{fig:theorem_drawing}
\end{figure}

\begin{remark}\label{remark:theorem formal}
    Consider $h(\cdot)$ is parameterized on the functional space $\mathcal{F}$, recall the definition of ``directional local minimum'' from \Cref{def:local_min}, \Cref{thm:formal} claims if the optimization finds $h(\cdot)\in\mathcal{F}_{inj}(g)$ and it satisfies the above conditions, then any perturbation from $\mathcal{F}_{skewsym\text{-}inj}(g)$ results in an increase of the loss function. \Cref{fig:theorem_drawing} provides an illustrative diagram of such relationship. In practice, once the gradient descent optimizer finds a function injective $h(\cdot)$ to $g(\cdot)$, the optimizer will stay inside $\mathcal{F}_{inj}(g)$, as the gradient descent method reaches a local minimum point for any perturbation step along $\mathcal{F}_{skewsym\text{-}inj}(g)$.

\end{remark}

As the major theoretical result of this paper, \Cref{thm:formal} narrates that through our proposed method, without knowing the exact invariant mapping function of the physics systems, the neural network is capable to recover it from the state neighborhood relationship in the image space. To the best of our knowledge, this is the first theoretical result on automated conservation law recovery, potentially building the bridge between AI and human-reliant scientific discovery work. Although the framework is used for dynamical system trajectory observations in this paper, the insight can be further extended to other physics or non-physics systems with inherent invariant properties.

%% file: 5_experiment.tex
\section{Experiments}\label{sec:experiments}
We first use two simple conservation examples to illustrate the ConCerNet procedures, then we discuss the model performance for a system with multiple conservation laws. In the end, we demonstrate the power of our method and extend the ConCerNet pipeline to high-dimensional problems by leveraging an autoencoder. As our task focuses on improving dynamical simulation trustworthiness on conservation properties, the conservation learning experiments are recorded in \Cref{tab:experiment_r_value} and the dynamical model performance is recorded in \Cref{tab:experiment_results}. We mainly compare ConCerNet with a baseline neural network, which does not include the projection module but shares the exact dynamics network structure and learning scheme/loss function with ConCerNet. We also compare ConCerNet with one classical modeling method (SINDy, \cite{brunton2016sindy}) and a DNN-based prior work (HNN, \cite{greydanus2019hamiltonian}) and delay the results to \Cref{appendix:experiment hnn}, where ConCerNet shows similar performance but ConCerNet is more generally applicable. All the experiments are performed over 5 random seeds, more system and experiment details are listed in \Cref{appendix:experiment details}.

\subsection{Simple Conservation Examples}
\begin{figure}[htb!]
\centering
\includegraphics[height=2.3in]{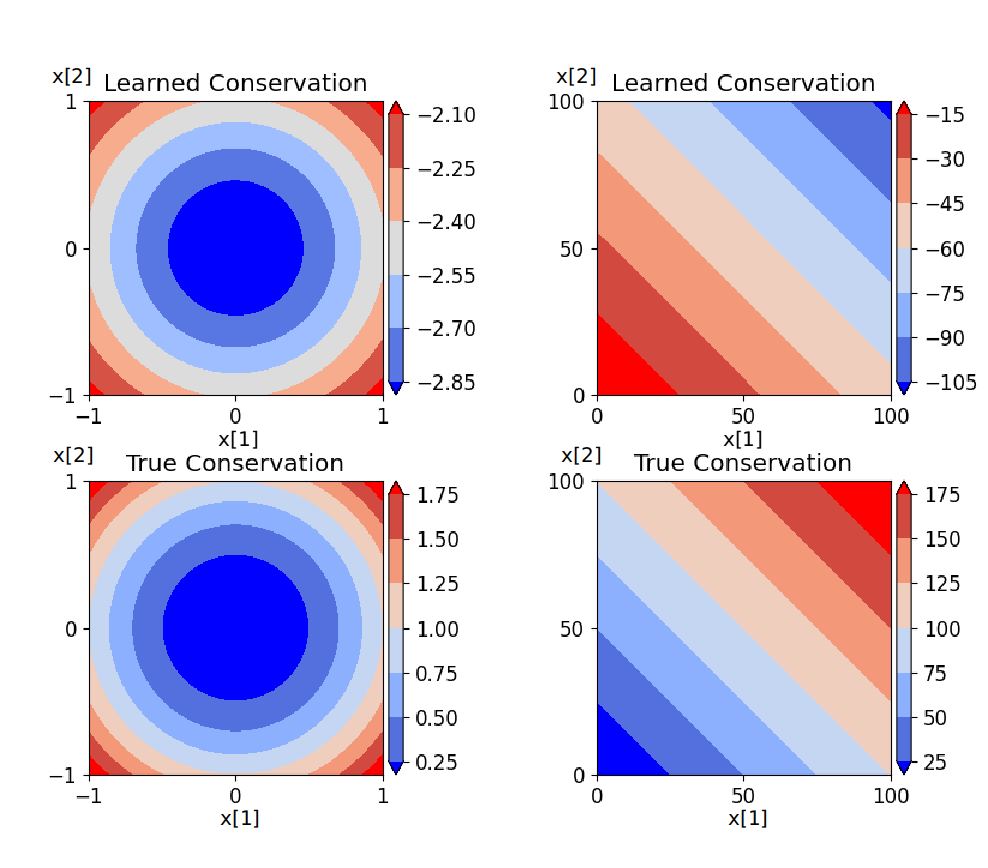}
\caption{Learned conservation function vs ground truth, left: ideal spring mass, right: chemical kinetics}
\label{fig:hx_springmass_and_chemicalreaction}
\end{figure}

\begin{figure}[htb!]
\centering
\includegraphics[width=0.7\columnwidth]{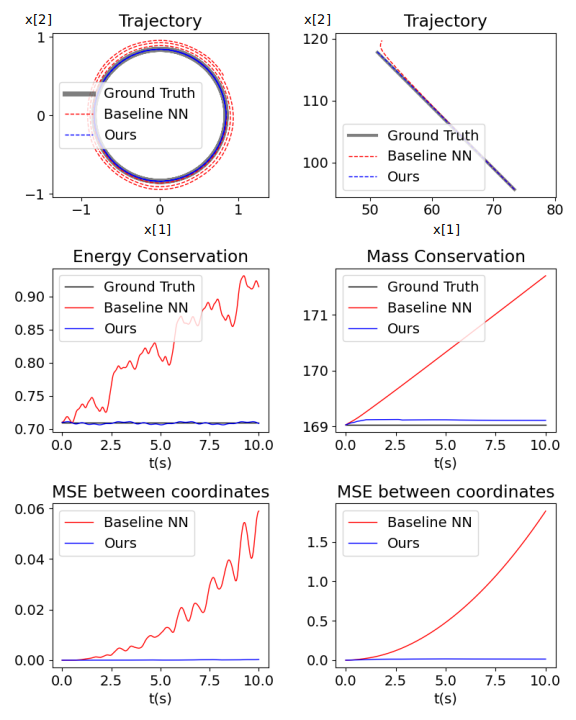}
\caption{Simulation comparisons of two simple examples: left column: ideal spring mass system, right column: chemical kinematics. 1st row: state trajectories, 2nd row: violation of conservation laws to ground truth, 3rd row: mean square error to ground truth. }
\label{fig:simulation_springmass_and_chemicalreaction}
\end{figure}

In this section, we introduce two simple examples: \textbf{Ideal spring mass system} under energy conservation ($x[1]^2+x[2]^2$) and \textbf{Chemical reaction} under mass conservation ($x[1]+x[2]$). Both systems have 2D state space for easier visualization of the learned conservation function. \Cref{fig:hx_springmass_and_chemicalreaction} shows the learned conservation compared with ground truth. The contrastive learning process captures the quadratic and linear functions, as the contour lines are drawn in circles and lines. To notice, the learned conservation function here is approximately the exact conservation function differing by some constant coefficient (we will further discuss their relationship in \Cref{discussion:learned_invariant}). This is a natural result because the SRL is invariant to linear transform of $H_{\theta_c}(\cdot)$; furthermore, the real-world conservation quantity is also a relative value instead of an absolute value. In \Cref{fig:simulation_springmass_and_chemicalreaction}, we compare the two methods by showing the typical trajectory, conservation violation and coordinate error to the ground truth. The vanilla neural network is likely to quickly diverge from the conserved trajectory, and the error grows faster than our proposed method.

\subsection{Systems with More Than One Conservation Functions}
In this section, we tackle a more complex system with more than one conservation law. The $\textbf{Kepler system}$ describes a planet orbiting around a star with elliptical trajectories. The planet's state is four-dimensional, including both coordinates in the 2D plane and the corresponding velocity. The system has two conservation terms (energy conservation $g_1(x)=\frac{x[3]^2+x[4]^2}{2}-\frac{1}{\sqrt{x[1]^2+x[2]^2}}$ and angular momentum conservation $g_2(x)=x[1]x[4]-x[2]x[3]$, we neglect the Laplace–Runge–Lenz vector for simplicity in this paper).  

In our prior discussion (\Cref{remark:H dim}), we claim that if the system has more than 1 conservation property, our method does not guarantee to find the injective function of each individual conservation equation. Consider the Kepler example, any functional combination of $g_1(x)$ and $g_2(x)$ is also conserved. In practice, we found the learned function is likely to converge to some linear function of the simpler conservation function. As we cannot visualize functions with four dimension input, we randomly sample 10 trajectories and calculate the $R^2$ by linear regressing the learned function towards each conservation function. The results in \Cref{tab:experiment_r_value} indicate that the learned function correlates with the angular momentum equation better than the energy equation which is more difficult to represent by the neural network. We also test the latent neural network with two-dimensional output ($m=2$), the two outputs are likely to be linear to each other, which does not affect dynamical prediction or the square ratio loss. Despite not learning all the  conservation laws, in the simulation stage, our method still outperforms the vanilla neural network by a large margin in both metrics with the learned invariant, as the results are shown in \Cref{tab:experiment_results}. 

\input{tables/experiments_r_value.tex}

\subsection{Larger System: Heat Equation}
To further extend our model to larger systems, we test our method on the \textbf{Heat Equation} on a 1D rod. The 1D rod is given some initial temperature distribution and insulated boundary conditions on both ends. The temperature $U(y,t)$, as a function of coordinate and time, gradually evens up following the heat equation $\frac{\partial U}{\partial t}=\frac{\partial^2 U}{\partial y^2}$. The total internal energy along the rod does not vary because the heat flow is blocked by the boundary. We use system state $x$ consisting of overall 101 nodes to discretize the entire interval $[-5,5]$ and compress system states to a 9 dimension latent space with an autoencoder pair $(E_{\theta_{E}},D_{\theta_{D}})$. For both contrastive conservation learning and dynamical system learning, original space state and time derivative $(x,\dot{x})$ are mapped to the autoencoder latent space $(z,\dot{z})$ by 
\begin{align}
    z & = E_{\theta_{E}} (x), 
    \dot{z}   = \frac{\partial E_{\theta_{E}}(x) } {\partial x } \times  \dot{x},
\end{align}
where $\times$ denotes the matrix multiplication by chain rule, the partial derivative from latent space to original space can be calculated by auto-differentiation package. After simulation, the latent space trajectory will be mapped back to the original space by $D_{\theta_{D}}$.

\begin{figure}[htb!]
\centering
\includegraphics[width=0.8\columnwidth]{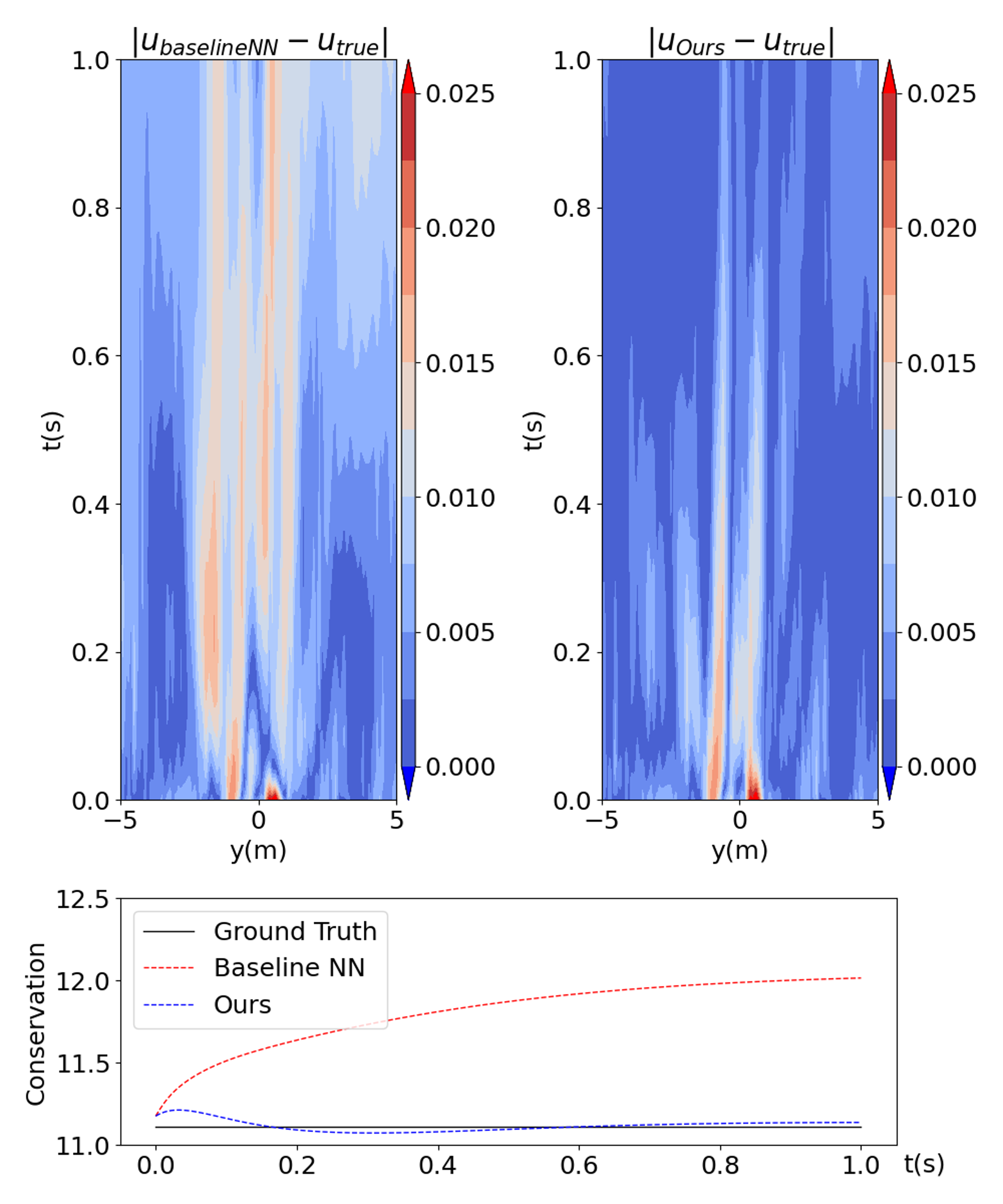}
\caption{Heat Equation Simulation. Upper: Simulation coordinate error comparison: vanilla neural network and ConCerNet. Bottom: Conservation violation to ground truth.}
\label{fig:heat_equation_cbombine}
\end{figure}

\Cref{tab:experiment_r_value} shows the neural network is capable to capture an affine function of mass conservation, regardless  
 of the reduced order and non-linear representation space. \Cref{fig:heat_equation_cbombine} draws the simulation result and conservation metric comparison between the vanilla neural network and our method. For both methods, the initial conservation violation error was introduced by the autoencoder. In general, our method conforms to ground truth trajectory and conservation laws much better than the vanilla method. 
 
\subsection{Experiment Summary}
\input{tables/experiment_results}

We generalize the quantitative results for all the experiments above in \Cref{tab:experiment_r_value} and \Cref{tab:experiment_results}. For conservation learning testing, we randomly sample 10 trajectories and linear regress the learned conservation function into each of the exact conservation laws and report the $R^2$ values. To test dynamics modeling, we sample 10 initial points and integrate the simulated trajectory with Runge-Kutta method, and calculate the trajectory state coordinate error and conservation violation to the ground truth. Our method outperforms the baseline neural networks as the error is often multiple times smaller. One may notice that the standard deviation is comparable with the error metric in the dynamics experiments, this is due to the instability of the dynamical systems. The trajectory tracking error will exponentially grow as a function of time, and the cases with outlier initialization are likely to dominate the averaged results and lead to large variance. In practice, our method is capable to control the tracking deviation better than the baseline method across almost all the cases.

%% file: tables/experiments_r_value.tex
\begin{table*}[htb!]
\vskip -0.1in
\centering
\caption{Average $R^2$ with linear regression between the learned function and exact conservations}
\scalebox{0.8}{
\begin{tabular}{c|c |c}
        \hline
        Task & Conservation & $R^2$ \\
        \hline
        Ideal spring mass system & $x[1]^2+x[2]^2=C$ & $0.997\pm 0.001$\\ \hline
        Chemical kinematics & $x[1]+x[2]=C$ & $0.999998\pm 1.9e-6$ \\\hline
        Kepler system & $\frac{x[3]^2+x[4]^2}{2}-\frac{1}{\sqrt{x[1]^2+x[2]^2}}=C_1$& $0.858\pm 0.104$\\
         & $x[1]x[4]-x[2]x[3]=C_2$ & $0.994\pm 0.001$\\ \hline
        Heat equation & $\int U dx = C$ & $0.9995\pm 0.0003$\\
        \hline
\end{tabular}}
\label{tab:experiment_r_value}
\end{table*}

%% file: tables/experiment_results.tex
\begin{table*}[htb!]
\vskip -0.1in
\centering
\caption{Simulation error over the tasks}
\scalebox{0.8}{
\begin{tabular}{c|c c c c}
        \hline
        \multirow{2}{*}{Task} & \multicolumn{2}{c}{Mean square error} &\multicolumn{2}{c}{Violation of conservation laws}   \\
        & Baseline NN  & ConCerNet & Baseline NN  & ConCerNet \\
        \hline
        Ideal spring mass system &0.022 $\pm$ 0.023 & \textbf{9.2e-3 $\pm$ 5.6e-3} & 0.012 $\pm$ 0.032 &  \textbf{1.4e-4 $\pm$ 7.4e-5}\\
        Chemical reaction &9.4e-3 $\pm$ 5.8e-3 & \textbf{5.9e-3$\pm$ 3.7e-3} & 0.019 $\pm$ 0.026 &  \textbf{5.6e-3 $\pm$ 5.4e-3} \\
        Kepler system & 0.794 $\pm$ 0.476 & \textbf{0.140 $\pm$ 0.067} & 0.016 $\pm$ 0.011 &  \textbf{5.2e-3 $\pm$ 4.2e-3}  \\
        Heat equation & 7.34e-5 $\pm$ 2.62e-5 & \textbf{2.79e-5 $\pm$ 1.28e-5} & 0.094 $\pm$ 0.070 &  \textbf{0.013 $\pm$ 0.008} \\
        \hline
\end{tabular}}
\vskip -0.1in
\label{tab:experiment_results}
\end{table*}

%% file: 6_discussion.tex
\section{Discussions}

\subsection{Learned Invariants vs. Exact Conservation laws}\label{discussion:learned_invariant}
From the theoretical side, for systems with only one conservation law, we prove through \Cref{thm:formal} that the optimizer converges to a composition function of the conservation function $g(\cdot)$. For systems with more than one conservation laws, it is difficult for the system to retrieve even one of the conservation equations, as any combination of individual conservation functions is also conserved, and \Cref{thm:formal} only guarantees the convergence to one of these combinations. Some prior work claiming to find the exact conservation functions depend on the prior human knowledge of conservation functions (i.e. using symbolic regression or limiting the fitting function classes). However, our paper takes a generic route using neural network parameterization to represent general functions, and therefore it can be extended to more complex systems or even non-linear representations like the heat equation case. In this paper, we do not involve symbolic regression based work in conservation learning comparison baselines as they are dependent on application-specific knowledge.

In practice, we found $h(\cdot)$ is likely to be a linear function of $g(\cdot)$, as \Cref{tab:experiment_r_value} shows the $R_2$ coefficients under linear regression, we can also tell the linear relationship from the contours in \Cref{fig:hx_springmass_and_chemicalreaction}. 



Despite the interesting experimental finding on the linear relationship, we yet have any theoretical ground for it. Our intuitive explanation is that the linear function is relatively easy for the neural network to express. We would like to emphasize that our paper focuses on improving the conservation performance for dynamical modeling, and we leave the task of finding the exact conservation laws or proving convergence to linear functions to future work. Regardless of the mapping between the learned function and the exact conservation function, the simulation is guaranteed to preserve the conservation property if the mapping is injective.

\subsection{ConCerNet Robustness to Noisy Observation and Contrastive Learning Data Efficiency}

To test the overall ConCerNet method performance under noisy observations, we perform experiments with different noise settings and compare the simulation error. The result is recorded in \Cref{tab:experiment_noise} in appendix. In general, ConCerNet is much less vulnerable to observation noise compared with baseline neural networks, as the projection layer provides a robust guarantee that the system will not deviate far away from the conservation manifold.

To test the contrastive conservation learning data efficiency, we conduct experiments with different observed trajectory numbers and points per trajectory and record the $R^2$ coefficient between the learned function and exact conservation law in \Cref{tab:data_efficiency_r_value} in \Cref{appendix:data efficiency}. In general, we usually need 100 to 1000 points to reach $R^2\approx1$ for the above four examples, the contrastive conservation learning is relatively data efficient.


\section{Conclusion}

In this paper, we propose ConCerNet, a generic framework to learn the dynamical system with designed features to preserve the invariant properties along the simulation trajectory. We first learn the conservation manifold in the state space from a contrastive learning perspective, then purposely enforce the dynamical system in the desired subspace leveraging a projection module. We establish the theoretical bridge between the learned latent representation and the actual conservation laws and experimentally validate the advantage of ConCerNet in both simulation error and conservation metrics and the extensibility to complex systems. Despite the paper presenting an end-to-end approach, both contrastive learning on system invariants and  projected dynamical system learning can be seen as an independent procedure and open up a different direction. We believe these ideas represent a promising route in automated invariant property discovery and practical dynamics modeling. 

%% file: 7_appendix.tex
\newpage
\appendix
\onecolumn
\section{Supplementary proofs}
\allowdisplaybreaks
\subsection{Proof of \Cref{prop:vectorspace}}
\Cref{prop:vectorspace} contains many claims, and we prove them one by one in the following. Recall $\mathcal{F}$ be the space of square integrable real-valued functions defined on a compact and convex set $\mathcal{X}\subset \mathbb{R}^n$ with positive volume.

\begin{enumerate}

    \item
    $\mathcal{F}$ is a vector space over the field $\mathbb{R}$.
    \begin{proof}
    $\quad$\\
    It is obvious the elements in $\mathcal{F}$ and scalars in $\mathbb{R}$ satisfy the associativity and commutativity of vector addition and distributivity of scalar multiplication. \\
    Closure of vector addition: $\forall f_1(\cdot), f_2(\cdot)\in \mathcal{F}, f_1(\cdot)+f_2(\cdot)\in \mathcal{F}$, because square integrability is preserved under finite summation.\\
    Zero element of vector addition: there exists zero vector $z(\cdot)\equiv0, z(\cdot)\in \mathcal{F}$, s.t. $\forall f(\cdot)\in \mathcal{F},z(\cdot)+f(\cdot)=f(\cdot)$.\\
    Inverse elements of vector addition: $\forall f(\cdot)\in \mathcal{F}$, there exists $(-f(\cdot))\in \mathcal{F},(-f(\cdot))+f(\cdot)=0$. \\
    Compatibility of scalar multiplication with field multiplication: $\forall a,b\in \mathbb{R}, f(\cdot)\in \mathcal{F}, a(bf(\cdot))=(ab)f(\cdot)$.\\
    Identity element of scalar multiplication: scalar $1\in\mathbb{R}$, $\forall f(\cdot)\in \mathcal{F}, 1f(\cdot)=f(\cdot)$.
    $\mathcal{F}$ and $\mathbb{R}$ satisfy the axioms of vector space, this completes the proof.
    \end{proof}

    \item 
    $\forall f_1(\cdot)\in\mathcal{F}$, $\mathcal{F}_{inj}(f_1)$ is a vector space over the field $\mathbb{R}$. 
    \begin{proof}
        $\quad$\\
        Pick two elements $f_{2,a}(\cdot), f_{2,b}(\cdot)$ from $\mathcal{F}_{inj}(f_1)$,
        by \Cref{def:injective}, $\forall x_i,x_j \in \mathcal{X}, f_1(x_i)=f_1(x_j) \implies f_{2,a}(x_i)=f_{2,a}(x_j),f_{2,b}(x_i)=f_{2,b}(x_j)$. Then we have $(f_{2,b}+f_{2,a})(x_i)=(f_{2,b}+f_{2,a})(x_j)\implies (f_{2,b}+f_{2,a})(\cdot)\in \mathcal{F}_{inj}(f_1)$, vector addition is closed.\\
        The other axioms are obvious.
    \end{proof}
    \item
        $\forall f_1(\cdot)\in\mathcal{F}$, $\mathcal{F}_{skewsym\text{-}inj}(f_1)$ is a vector space over the field $\mathbb{R}$.
    \begin{proof}
        $\quad$\\
       
        Pick two elements $f_{2,a}(\cdot), f_{2,b}(\cdot)$ from $\mathcal{F}_{skewsym\text{-}inj}(f_1)$,
        by \Cref{def:injective}, $f_1(\cdot)$ if $\forall r\in\mathbb{R}, \mathbb{E}[ f_{2,a}(x)|f_1(x)=r]=0,\mathbb{E}[ f_{2,b}(x)|f_1(x)=r]=0$. Then we have $\mathbb{E}[ f_{2,a}(x)+f_{2,b}(x)|f_1(x)=r]=0\implies (f_{2,b}+f_{2,a})(\cdot)\in \mathcal{F}_{skewsym\text{-}inj}(f_1)$, vector addition is closed.\\
        The other axioms are obvious.
    \end{proof}

    \item 
    $\forall f_1(\cdot)\in\mathcal{F}$, $\mathcal{F}_{skewsym\text{-}inj}(f_1)\cap \mathcal{F}_{inj}(f_1)=\{\text{zero vector}\}$, where the zero vector here is $z(\cdot)\equiv0$.
    \begin{proof}
        $\quad$\\
        We prove the claim by contradiction. Suppose there exists $x_0\in\mathcal{X}$ and non-zero $f_2(\cdot)\in \mathcal{F}_{skewsym\text{-}inj}(f_1)\cap \mathcal{F}_{inj}(f_1),f_2(x_0)=s\neq 0$. Because $f_2(\cdot)\in \mathcal{F}_{inj}(f_1)$, $\forall f_1(x)=f_1(x_0), f_2(x)=s$, then we have $\mathbb{E}[f_2(x)|f_1(x)=f_1(x_0)]=s\mathbb{E}[1|f_1(x)=f_1(x_0)]=s\neq0$. This contradicts with the definition of $ \mathcal{F}_{skewsym\text{-}inj}(f_1)$ and completes the proof.
    \end{proof}

    \item 
        $\forall f_1(\cdot)\in\mathcal{F}$, $\mathcal{F}_{skewsym\text{-}inj}(f_1)\cup \mathcal{F}_{inj}(f_1)=\mathcal{F}$.
    \begin{proof}
        $\quad$\\
        For arbitrary $f_1(\cdot),f_2(\cdot)\in \mathcal{F}$, let $f_{2,inj}(x)=\mathbb{E}_{x'}[f_2(x')|f_1(x')=f_1(x)], f_{2,skewsym\text{-}inj}(x)=f_2(x)-f_{2,inj}(x)$. \\
        For $f_{2,inj}(x)$, $\forall x_i,x_j\in\mathcal{X}, \text{ if } f_1(x_i)=f_1(x_j)$,

        \begin{align*}
            f_{2,inj}(x_i) & =\mathbb{E}_{x'}[f_2(x')|f_1(x')=f_1(x_i)]\\
            & = \mathbb{E}_{x'}[f_2(x')|f_1(x')=f_1(x_j)]\\
            &=f_{2,inj}(x_j)
        \end{align*}
        Next, we need to prove the integrability of $f_{2,inj}(\cdot)$ and $f_{2,skewsym\text{-}inj}(\cdot)$. In \Cref{prop:vectorspace}, we assume $f_1(\cdot)$ and $f_2(\cdot)$ are square integrable on the c set $\mathcal{X}$. Square integrability on compact set indicates function boundedness (i.e. $\exists C > 0, \forall x \in \mathcal{X}, |f_1(x)|,|f_2(x)|<C$).

        $f_{2,inj}(x)=\mathbb{E}_{x'}[f_2(x')|f_1(x')=f_1(x)]$ is the expectation of a square integrable function on a subset of $X$ ($f_1^{-1}(f_1(x))$), calculate the square integral of $f_{2,inj}(\cdot)$ on this subset:
        \begin{align*}
            \int_{x\in f_1^{-1}(f_1(x))} f_{2,inj}^2(x)dx & =\int_{x\in f_1^{-1}(f_1(x))} \mathbb{E}^2_{x'}[f_2(x')|f_1(x')=f_1(x)] dx\\
            & \leq \int_{x\in f_1^{-1}(f_1(x))} \mathbb{E}_{x'}[f^2_2(x')|f_1(x')=f_1(x)]dx\\
            & = \int_{x\in f_1^{-1}(f_1(x))} f^2_2(x)dx
        \end{align*}
        where the inequality comes from variance-mean relationship (i.e. $\mathbb{E}^2[x]=\mathbb{E}[x^2]-\text{Var}(x)$).

        For each subset of $X$, the square integral of $f_{2,inj}(\cdot)$ is less or equal to that of $f_{2}(\cdot)$. Then for the whole set $X$, we have 
        \begin{align*}
            \int_{x\in \mathcal{X}} f_{2,inj}^2(x)dx \leq \int_{x\in \mathcal{X}} f^2_2(x)dx 
        \end{align*}

        With $f_2(\cdot)\in\mathcal{F}$ being square integrable on $\mathcal{X}$,$f_{2,inj}(\cdot)$ is also square integrable on $\mathcal{X}$ (so as $f_{2,skewsym\text{-}inj}(\cdot)=f_2(\cdot)-f_{2,inj}(\cdot)$ because the square integrability is preserved under subtraction/sum). Therefore, we have $f_{2,inj}(\cdot)\in \mathcal{F}_{inj}(f_1)$.\\
        For $f_{2,skewsym\text{-}inj}(x)$, $\forall r\in\mathbb{R}$,
        \begin{align*}
            &\mathbb{E}_{x'}[f_{2,skewsym\text{-}inj}(x')|f_1(x')=r]  \\
            & =\mathbb{E}_{x'}[f_2(x')-\mathbb{E}_{x''}[f_2(x'')|f_1(x'')=r]|f_1(x')=r] \\
            & =\mathbb{E}_{x'}[f_2(x')|f_1(x')=r]-\mathbb{E}_{x''}[f_2(x'')|f_1(x'')=r] \\
            & = 0
        \end{align*}
        Therefore, we have $f_{2,skewsym\text{-}inj}\in \mathcal{F}_{skewsym\text{-}inj}(f_1)$. $\forall f_1(\cdot), f_2(\cdot)\in \mathcal{F}$, it can be decomposed into $f_{2,skewsym\text{-}inj}\in \mathcal{F}_{skewsym\text{-}inj}(f_1)$ and $f_{2,inj}\in\mathcal{F}_{inj}(f_1)$.
    \end{proof}
\end{enumerate}

\subsection{Justification of \Cref{ass:epsilon}}\label{appendix:epsilonjustify}

In the \Cref{thm:formal}, we assume the found $h(\cdot)$ is functional injective and relative continuous to $g(\cdot)$, the states within the preimage of $g(\cdot)$ will have similar values under the $h(\cdot)$ function mapping as well. 
 Then we can also approximate $h(\cdot)$ with Taylor expansion in $g^{-1}_{\mathcal{X}}(g(x),\epsilon)$ and higher-order terms can be neglected when $\epsilon$ is small. The following approximation is for 1D scenario but can be extended to multi-dimension $x$. We also only consider single neighborhood cases, if there exist multiple neighborhoods, we can take the supreme as the upper bound of the LHS in \Cref{ass:epsilon}.

When $\epsilon$ is small, the preimage including $x$ can be estimated by $g^{-1}_{\mathcal{X}}(g(x),\epsilon)\approx [x-\epsilon/(g'(x)),x+\epsilon/(g'(x))]$, where $g'(x)$ denotes the derivative of $g(x)$.

\begin{align*}
    & \int_{x'\in g^{-1}_{\mathcal{X}}(g(x),\epsilon)}(h(x)-h(x'))dx' 
    \approx \int_{x-\epsilon/g'(x)}^{x+\epsilon/g'(x)}(h(x)-h(x'))dx' \\
    \shortintertext{we need to consider 2nd order term because the first order integrates to 0}
    & \approx \int_{x-\epsilon/g'(x)}^{x+\epsilon/g'(x)} (h'(x)(x'-x)+\frac{1}{2}h''(x)(x'-x)^2)dx' \\
    & = [h'(x)\xi^2+\frac{1}{6}h''(x)\xi^3]\bigg\rvert_{\xi = -\epsilon/g'(x)}^{\xi = +\epsilon/g'(x)}
    = \frac{1}{3}\frac{h''(x)\epsilon^3}{g'(x)^3}
\end{align*}
\begin{align*}
    & \int_{x'\in g^{-1}_{\mathcal{X}}(g(x),\epsilon)}(h(x)-h(x'))^2dx' \\
     \shortintertext{we can drop the 2nd order term}
    & \approx \int_{x-\epsilon/g'(x)}^{x+\epsilon/g'(x)} (h'(x)(x'-x))^2dx' 
    = [h''(x)\xi^3]\bigg\rvert_{\xi = -\epsilon/g'(x)}^{\xi = +\epsilon/g'(x)}\\
    & = \frac{2}{3}\frac{(h'(x))^2\epsilon^3}{g'(x)^3}\\
\end{align*}
\begin{align*}
    & \frac{\int_{x'\in g^{-1}_{\mathcal{X}}(g(x),\epsilon)}(h(x)-h(x'))dx'}{{\int_{x'\in g^{-1}_{\mathcal{X}}(g(x),\epsilon)}(h(x)-h(x'))^2}dx'}
     \approx \frac{1}{3}\frac{h''(x)\epsilon^3}{g'(x)^3} /  \frac{2}{3}\frac{(h'(x))^2\epsilon^3}{g'(x)^3}\\
    & = \frac{1}{2}\frac{h''(x)}{h'(x)^2} \leq \max_{x\in\mathcal{X}} \frac{1}{2}\frac{h''(x)}{h'(x)^2} = \text{Constant}
\end{align*}
Then we have $LHS\approx C_1\epsilon \cdot\text{ Constant}\rightarrow 0$ when $\epsilon\rightarrow0$, therefore, there always exists a small $\epsilon$ to satisfy this condition.

Notice the above approximation and constant upper bound implicitly lie on that $|g'(x)|\neq0$ (thus $|h'(x)|\neq 0$ by relative Lipschitz continuity in \Cref{ass:rela_continuous}). When these derivatives are 0, we can use high order derivatives to replace them (Higher order non-zero derivative always exists as $g(\cdot)$ is non-constant and smooth).

\subsection{Proof of \Cref{thm:formal}}\label{appendix:proof}

\begin{proof}
    Let the perturbed $h(x)=\lim_{\delta \to 0} h_1(x)+\delta h_2(x)$, where $h_1(\cdot) \in F_{inj}(g)$ and $h_2 (\cdot)\in F_{skewsym\text{-}inj}(g)$. Let $\mathcal{L}$ denote the integral square ratio loss $\mathcal{L}_{ISQL}$ for notation simplicity. For arbitrary fixed $h_1(\cdot), h_2(\cdot)$ and small $\epsilon$, consider $\mathcal{L}$ being a function of $\delta$: $\mathcal{L}(h(\cdot),\epsilon)=\mathcal{L}(\delta)$, our proof goal is $\frac{dL}{d\delta}=0$ and $\frac{d^2L}{d\delta^2}>0$. Hence, we can claim if we found $h(\cdot)=h_1(\cdot)\in F_{inj}(g)$ and satisfies \Cref{ass:rela_continuous}, \Cref{ass:gen_variance} and \ref{ass:epsilon}, there exists $\overline{\delta}>0, s.t. \forall \delta < \overline{\delta}, h_2(\cdot)\in\mathcal{F}_{skewsym\text{-}inj}, \mathcal{L}(h_1(\cdot)+\delta h_2(\cdot))>\mathcal{L}(h_1(\cdot))$. Then $h(\cdot)$ is a directional local minimum point against any perturbation on $F_{skewsym\text{-}inj}(g)$. We prove the existence of $\overline{\delta}$ by taking $\delta \rightarrow 0$.

    By \Cref{def:injective}, if $h_2(\cdot) \in F_{skewsym\text{-}inj}(g)$, $\forall r \in \mathbb{R}$, we have $\mathbb{E}_{x}[h_2(x)|g(x)=r]=0$. Suppose we only perturb $h(\cdot)$ by $\delta$ at single interval where $g(x)=r$, this will violate the property of $h_2(\cdot)$, because the expectation of $h_2(\cdot)$ on $g^{-1}(r)$ will also be non-zero. To compensate the violation, we need to simultaneously perturb $h(\cdot)$ on other intervals within $g^{-1}(r)$, subject to $\mathbb{E}_{x}[h_2(x)|g(x)=r]=0$. In optimal control theory \cite{liberzon2011calculus}, to include a larger family of control perturbations (e.g. discontinuous control input) and maintain the constraint of the right endpoint, a typical approach is to add two ``needle'' perturbations, where the control input is perturbed at two intervals. Inspired by this approach, we draw two non-overlapping intervals with the same volume from the preimage $g^{-1}(r)$, and perturb them by an inverse pair of magnitude. This symmetric perturbation function belongs to $F_{skewsym\text{-}inj}(g)$ as its expectation on $g^{-1}(r)$ is always zero. This ``needle'' pair can be viewed as a basic perturbation in $F_{skewsym\text{-}inj}(g)$ and any $h_2(\cdot)\in F_{skewsym\text{-}inj}(g)$ can be decomposed into these pairs.

    We formulate this perturbation as following: for certain pre-image $g^{-1}(r)$, we choose two intervals $\mathcal{X}_i,\mathcal{X}_j\subset g^{-1}(r), \int_{x\in\mathcal{X}_i}1dx = \int_{x\in\mathcal{X}_i}1dx=\tau>0,\mathcal{X}_i\cap \mathcal{X}_j=\emptyset$. Because we will append another infinitesimal coefficient $\delta$ ahead of $h_2(\cdot)$, the absolute magnitude of the perturbation function (or the ``needle height'') does not matter. Without loss of generality, let the needle function be:
    \begin{equation*}
        h_2(x)= 
        \begin{cases}
        +1 & \text{if } x \in  \mathcal{X}_i \\
         -1  & \text{if } x \in \mathcal{X}_j\\
        0 & \text{otherwise.}
        \end{cases}
    \end{equation*}


To simplify notation, let $h_1(x)=s,\forall x\in g^{-1}(r)$, $g^{-1}(g(x_i),\epsilon)=\overline{g}$. Then the perturbed function is
    \begin{equation*}
        h(x)= h_1(x)+\delta h_2(x) = 
        \begin{cases}
        h_1(x)+ \delta = s+\delta & \text{if } x \in\mathcal{X}_i\\
        h_1(x)- \delta= s-\delta  & \text{if } x \in\mathcal{X}_j\\
        h_1(x)& \text{otherwise.}
    \end{cases}
    \end{equation*}   

\begin{align*}
    \mathcal{L}& = \mathcal{L}(h_1)  =  \int_{x\in\mathcal{X}}\frac{\int_{x'\in \overline{g}}(h_1(x)-h_1(x'))^2dx'}{\int_{x''\in \mathcal{X}}(h_1(x)-h_1(x''))^2dx''}dx \\
    & = \int_{x\in\mathcal{X}\backslash \overline{g}}\frac{\int_{x'\in \overline{g}}(h_1(x)-h_1(x'))^2dx'}{\int_{x''\in \mathcal{X}}(h_1(x)-h_1(x''))^2dx''}dx + \int_{ \overline{g}\backslash \{\mathcal{X}_i,\mathcal{X}_j\} }\frac{\int_{x'\in \overline{g}}(h_1(x)-h_1(x'))^2dx'}{\int_{x''\in \mathcal{X}}(h_1(x)-h_1(x''))^2dx''}dx \\
    & + \int_{x\in\mathcal{X}_i}\frac{\int_{x'\in \overline{g}}(h_1(x)-h_1(x'))^2dx'}{\int_{x''\in \mathcal{X}}(h_1(x)-h_1(x''))^2dx''}dx + \int_{x\in\mathcal{X}_j}\frac{\int_{x'\in \overline{g}}(h_1(x)-h_1(x'))^2dx'}{\int_{x''\in \mathcal{X}}(h_1(x)-h_1(x''))^2dx''} dx
\end{align*}

\begin{align*}
    \mathcal{L} +\delta \mathcal{L} 
    & =  \mathcal{L}(h) =  \mathcal{L}(h_1+\delta h_2)\\
    & = \int_{x\in\mathcal{X}\backslash \overline{g}}\frac{\int_{x'\in \overline{g}}(h(x)-h(x'))^2dx'}{\int_{x''\in \mathcal{X}}(h(x)-h(x''))^2dx''}dx + \int_{ \overline{g}\backslash \{\mathcal{X}_i,\mathcal{X}_j\} }\frac{\int_{x'\in \overline{g}}(h(x)-h(x'))^2dx'}{\int_{x''\in \mathcal{X}}(h(x)-h(x''))^2dx''}dx \\
    & + \int_{x\in\mathcal{X}_i}\frac{\int_{x'\in \overline{g}}(h(x)-h(x'))^2dx'}{\int_{x''\in \mathcal{X}}(h(x)-h(x''))^2dx''}dx + \int_{x\in\mathcal{X}_j}\frac{\int_{x'\in \overline{g}}(h(x)-h(x'))^2dx'}{\int_{x''\in \mathcal{X}}(h(x)-h(x''))^2dx''} dx
\end{align*}

Let $\overline{\mathcal{L}}_1,\overline{\mathcal{L}}_2,\overline{\mathcal{L}}_3,\overline{\mathcal{L}}_4$ denote the 4 terms in the above bracket, address each term:
    
\begin{align*}
    &\overline{\mathcal{L}}_1 + \delta \overline{\mathcal{L}}_1 \\
    & = \int_{x\in\mathcal{X}\backslash \overline{g}}\frac{\int_{x'\in \{\mathcal{X}_i,\mathcal{X}_j\}}(h(x)-h(x'))^2dx'+\int_{x'\in \overline{g}\backslash \{\mathcal{X}_i,\mathcal{X}_j\}}(h(x)-h(x'))^2dx'}{\int_{x''\in  \{\mathcal{X}_i,\mathcal{X}_j\}}(h(x)-h(x''))^2dx''+\int_{x''\in \mathcal{X}\backslash \{\mathcal{X}_i,\mathcal{X}_j\}}(h(x)-h(x''))^2dx''}dx \\
    & = \int_{x\in\mathcal{X}\backslash \overline{g}}\frac{2\delta^2\tau+\int_{x'\in \overline{g}}(h_1(x)-h_1(x'))^2dx'}{\int_{x''\in  \{\mathcal{X}_i,\mathcal{X}_j\}}(h(x)-h(x''))^2dx''-\int_{x''\in \{\mathcal{X}_i,\mathcal{X}_j\}}(h(x)-h(x''))^2dx''+\int_{x''\in \mathcal{X}}(h(x)-h(x''))^2dx''}dx \\
    & =\int_{x\in\mathcal{X}\backslash \overline{g}}\frac{2\delta^2\tau+\int_{x'\in \overline{g}}(h_1(x)-h_1(x'))^2dx'}{[-2(h_1(x)-s)^2+(h_1(x)-s-\delta)^2+(h_1(x)-s+\delta)^2]\tau+\int_{x''\in \mathcal{X}}(h_1(x)-h_1(x''))^2dx''}dx \\
    & = \int_{x\in\mathcal{X}\backslash \overline{g}}\frac{2\delta^2\tau+\int_{x'\in \overline{g}}(h_1(x)-h_1(x'))^2dx'}{2\delta^2\tau+\int_{x''\in \mathcal{X}}(h_1(x)-h_1(x''))^2dx''}dx\\
    \shortintertext{neglect \scalebox{.7}{$\mathcal{O}$}$(\delta^2)$  terms}
    & \approx \int_{x\in\mathcal{X}\backslash \overline{g}}\frac{\int_{x'\in \overline{g}}(h_1(x)-h_1(x'))^2dx'}{\int_{x''\in \mathcal{X}}(h_1(x)-h_1(x''))^2dx''}[1-\frac{2\delta^2\tau}{\int_{x''\in \mathcal{X}}(h_1(x)-h_1(x''))^2dx''}+\frac{2\delta^2\tau}{\int_{x'\in \overline{g}}(h_1(x)-h_1(x'))^2dx''}]dx
\end{align*}
\begin{align*}
    &\overline{\mathcal{L}}_2 + \delta \overline{\mathcal{L}}_2 \\
    &=\int_{ x\in\overline{g}\backslash \{\mathcal{X}_i,\mathcal{X}_j\}}\frac{\int_{x'\in \{\mathcal{X}_i,\mathcal{X}_j\}}(h(x)-h(x'))^2dx'+\int_{x'\in \overline{g}\backslash \{\mathcal{X}_i,\mathcal{X}_j\}}(h(x)-h(x'))^2dx'}{\int_{x''\in \{\mathcal{X}_i,\mathcal{X}_j\}}(h(x)-h(x''))^2dx''+\int_{x''\in \mathcal{X}\backslash \{\mathcal{X}_i,\mathcal{X}_j\}}(h(x)-h(x''))^2dx''}dx\\
    & =\int_{x\in\overline{g}\backslash \{\mathcal{X}_i,\mathcal{X}_j\}}\frac{2\delta^2\tau+\int_{x'\in \overline{g}}(h_1(x)-h_1(x'))^2dx'}{2\delta^2\tau+\int_{x''\in \mathcal{X}}(h_1(x)-h_1(x''))^2dx''}dx \\
    \shortintertext{neglect \scalebox{.7}{$\mathcal{O}$}$(\delta^2)$  terms}
    & = \int_{x\in\overline{g}\backslash \{\mathcal{X}_i,\mathcal{X}_j\}}\frac{\int_{x'\in \overline{g}}(h_1(x)-h_1(x'))^2dx'}{\int_{x''\in \mathcal{X}}(h_1(x)-h_1(x''))^2dx''}[1-\frac{2\delta^2\tau}{\int_{x''\in \mathcal{X}}(h_1(x)-h_1(x''))^2dx''}+\frac{2\delta^2\tau}{\int_{x'\in \overline{g}}(h_1(x)-h_1(x'))^2dx''}]dx
\end{align*}

\begin{align*}
    \overline{\mathcal{L}}_3 + \delta \overline{\mathcal{L}}_3 & =\int_{x\in\mathcal{X}_i}\frac{\int_{x'\in \overline{g}}(h(x)-h(x'))^2dx'}{\int_{x''\in \mathcal{X}}(h(x)-h(x''))^2dx''}dx \\
    & =\int_{x\in\mathcal{X}_i} \frac{\int_{x'\in \overline{g}}(s+\delta-h(x'))^2dx'}{\int_{x''\in \mathcal{X}}(s+\delta-h(x''))^2dx''}dx\\
    \shortintertext{perturbation at $x\in\mathcal{X}_i$ will also affect $h(x')$ and $h(x'')$ terms, but this deviation will be canceled by the deviation of $h(x)$}
    & = \int_{x\in\mathcal{X}_i}\frac{\int_{x'\in \overline{g}\backslash \{\mathcal{X}_i,\mathcal{X}_j\}}[(s-h_1(x'))^2+2\delta(s-h_1(x'))+\delta^2]dx'+\int_{x'\in  \mathcal{X}_j}[(s+\delta-s+\delta)^2]dx'}{\int_{x''\in \mathcal{X}\backslash \{\mathcal{X}_i,\mathcal{X}_j\}}[(s-h_1(x''))^2+2\delta(s-h_1(x''))+\delta^2]dx''+\int_{x''\in  \mathcal{X}_j}[(s+\delta-s+\delta)^2]dx''}dx
    \shortintertext{using $h_1(x)=s,\forall x\in \mathcal{X}_i\bigcup \mathcal{X}_j\subset g^{-1}(r)$ and neglect higher order terms with $\lim_{a\to 0}\frac{1}{1+a}\approx 1-a+a^2$, let $|\mathcal{X}|=\int_{x\in\mathcal{X}}1dx,|\overline{g}|=\int_{x\in\overline{g}}1dx $ be the measure of $\mathcal{X}$ and $\overline{g}$}
    & \approx \int_{x\in\mathcal{X}_i} \frac{\int_{x'\in \overline{g}}(s-h_1(x'))^2dx'}{\int_{x''\in \mathcal{X}}(s-h_1(x''))^2dx''}*[1+\frac{2\delta\int_{x'\in \overline{g}}(s-h_1(x'))dx'+(|\overline{g}|+2\tau)\delta^2}{\int_{x'\in \overline{g}}(s-h_1(x'))^2dx'}]\\
    & *[1-\frac{2\delta\int_{x''\in \mathcal{X}}(s-h_1(x''))dx''+(|\mathcal{X}|+2\tau)\delta^2}{\int_{x''\in \mathcal{X}}(s-h_1(x''))^2dx''}+4\delta^2 {\frac{\int_{x''\in \mathcal{X}}(s-h_1(x''))dx''}{\int_{x''\in \mathcal{X}}(s-h_1(x''))^2dx''}}^2]dx \\
\end{align*}

Apply the same treatment to $\overline{\mathcal{L}}_4 + \delta \overline{\mathcal{L}}_4 $, notice adding two terms will cancel $\delta$ terms
\begin{align*}
    & \overline{\mathcal{L}}_3 + \delta \overline{\mathcal{L}}_3+\overline{\mathcal{L}}_4 + \delta \overline{\mathcal{L}}_4 \\
    & \approx  2\tau\frac{\int_{x'\in \overline{g}}(s-h_1(x'))^2dx'}{\int_{x''\in \mathcal{X}}(s-h_1(x''))^2dx''}*[1+\frac{(|\overline{g}|+2\tau)\delta^2}{\int_{x'\in \overline{g}}(s-h_1(x'))^2dx'}-\frac{(|\mathcal{X}|+2\tau)\delta^2}{\int_{x''\in \mathcal{X}}(s-h_1(x''))^2dx''}\\
    & +4\delta^2 {\frac{\int_{x''\in \mathcal{X}}(s-h_1(x''))dx''}{\int_{x''\in \mathcal{X}}(s-h_1(x''))^2dx''}}^2-4\delta^2 \frac{\int_{x''\in \mathcal{X}}(s-h_1(x''))dx''}{\int_{x''\in \mathcal{X}}(s-h_1(x''))^2dx''} \frac{\int_{x'\in \overline{g}}(s-h_1(x'))dx'}{\int_{x'\in \overline{g}}(s-h_1(x'))^2dx'}]\\
    & =  2\tau\frac{\int_{x'\in \overline{g}}(s-h_1(x'))^2dx'}{\int_{x''\in \mathcal{X}}(s-h_1(x''))^2dx''}*[1+\frac{|\overline{g}|\delta^2}{\int_{x'\in \overline{g}}(s-h_1(x'))^2dx'}-\frac{|\mathcal{X}|\delta^2}{\int_{x''\in \mathcal{X}}(s-h_1(x''))^2dx''}\\
    & +4\delta^2 {\frac{\int_{x''\in \mathcal{X}}(s-h_1(x''))dx''}{\int_{x''\in \mathcal{X}}(s-h_1(x''))^2dx''}}^2-4\delta^2 \frac{\int_{x''\in \mathcal{X}}(s-h_1(x''))dx''}{\int_{x''\in \mathcal{X}}(s-h_1(x''))^2dx''} \frac{\int_{x'\in \overline{g}}(s-h_1(x'))dx'}{\int_{x'\in \overline{g}}(s-h_1(x'))^2dx'}] \\
    & + \int_{x\in \{\mathcal{X}_i,\mathcal{X}_j\}}\frac{\int_{x'\in \overline{g}}(h_1(x)-h_1(x'))^2dx'}{\int_{x''\in \mathcal{X}}(h_1(x)-h_1(x''))^2dx''}[-\frac{2\delta^2\tau}{\int_{x''\in \mathcal{X}}(h_1(x)-h_1(x''))^2dx''}+\frac{2\delta^2\tau}{\int_{x'\in \overline{g}}(h_1(x)-h_1(x'))^2dx''}]
\end{align*}
sum up all terms:
\begin{align*}
    & \sum_{k=1}^4[\overline{\mathcal{L}}_k + \delta \overline{\mathcal{L}}_k]\\
    & = \int_{x\in\mathcal{X}}\frac{\int_{x'\in \overline{g}}(h_1(x)-h_1(x'))^2dx'}{\int_{x''\in \mathcal{X}}(h_1(x)-h_1(x''))^2dx''}dx - 2\delta^2\tau\int_{x\in\mathcal{X}}\frac{\int_{x'\in \overline{g}}(h_1(x)-h_1(x'))^2dx'}{\int_{x''\in \mathcal{X}}(h_1(x)-h_1(x''))^2dx''}*\frac{1}{\int_{x''\in \mathcal{X}}(h_1(x)-h_1(x''))^2}dx \\
    & + 2\delta^2\tau\int_{x\in\overline{g}}\frac{\int_{x'\in \overline{g}}(h_1(x)-h_1(x'))^2dx'}{\int_{x''\in \mathcal{X}}(h_1(x)-h_1(x''))^2dx''}*\frac{1}{\int_{x'\in \overline{g}}(h_1(x)-h_1(x'))^2dx'}dx\\
    &+2\frac{\int_{x'\in \overline{g}}(s-h_1(x'))^2dx'}{\int_{x''\in \mathcal{X}}(s-h_1(x''))^2dx''}*[\frac{|\overline{g}|\delta^2\tau}{\int_{x'\in \overline{g}}(s-h_1(x'))^2dx'}-\frac{|\mathcal{X}|\delta^2\tau}{\int_{x''\in \mathcal{X}}(s-h_1(x''))^2dx''}\\
    & +4\delta^2\tau {\frac{\int_{x''\in \mathcal{X}}(s-h_1(x''))dx''}{\int_{x''\in \mathcal{X}}(s-h_1(x''))^2dx''}}^2-4\delta^2\tau \frac{\int_{x''\in \mathcal{X}}(s-h_1(x''))dx''}{\int_{x''\in \mathcal{X}}(s-h_1(x''))^2dx''} \frac{\int_{x'\in \overline{g}}(s-h_1(x'))dx'}{\int_{x'\in \overline{g}}(s-h_1(x'))^2dx'}] \\
    & = \sum_{k=1}^4\overline{\mathcal{L}}_k  \\
    & \quad +2\delta^2\tau \Biggr[ -\int_{x\in\mathcal{X}}\frac{\int_{x'\in \overline{g}}(h_1(x)-h_1(x'))^2dx'}{\int_{x''\in \mathcal{X}}(h_1(x)-h_1(x''))^2dx''}*\frac{1}{\int_{x''\in \mathcal{X}}(h_1(x)-h_1(x''))^2dx''}dx \cdots\cdots\cdots\cdots\textcircled{1}  \\
    & \qquad  \qquad   + \int_{x\in\overline{g}}\frac{\int_{x'\in \overline{g}}(h_1(x)-h_1(x'))^2dx'}{\int_{x''\in \mathcal{X}}(h_1(x)-h_1(x''))^2dx''}*\frac{1}{\int_{x'\in \overline{g}}(h_1(x)-h_1(x'))^2dx'}dx\cdots\cdots\cdots\cdots\cdots\textcircled{2}\\
    &\qquad  \qquad  +\frac{\int_{x'\in \overline{g}}(s-h_1(x'))^2dx'}{\int_{x''\in \mathcal{X}}(s-h_1(x''))^2dx''}*[\frac{|\overline{g}|}{\int_{x'\in \overline{g}}(s-h_1(x'))^2dx'}-\frac{|\mathcal{X}|}{\int_{x''\in \mathcal{X}}(s-h_1(x''))^2dx''}\cdots\cdots\textcircled{3}\\
    & \qquad  \qquad  +4\frac{\int_{x''\in \mathcal{X}}(s-h_1(x''))dx''}{\int_{x''\in \mathcal{X}}(s-h_1(x''))^2dx''}(\frac{\int_{x''\in \mathcal{X}}(s-h_1(x''))dx''}{\int_{x''\in \mathcal{X}}(s-h_1(x''))^2dx''} - \frac{\int_{x'\in \overline{g}}(s-h_1(x'))dx'}{\int_{x'\in \overline{g}}(s-h_1(x'))^2dx'})] \Biggr] \cdots\cdots\textcircled{4} \\
\end{align*}

There is no first order $\delta$ term in the perturbed loss function ($\frac{d\mathcal{L}}{d\delta}=0$), then we proved $h(\cdot)$ is a stationary point to any perturbation from $F_{skewsym\text{-}inj}(g)$ .

Before we evaluate the second order coefficients inside the big bracket, we first derive two useful inequalities regarding the generalized variance:



Recall we assume $h_1(\cdot)$ is $C_1$ relative Lipschitz continuous to $g(\cdot)$, $\forall x\in\mathcal{X}_i\cap \mathcal{X}_j, x'\in g^{-1}_{\mathcal{X}} (r,\epsilon)$, then 
\begin{align*}
    (h_1(x)-h_1(x'))^2 & \leq C_1^2(g(x)-g(x'))^2 \\
    & \leq C_1^2 \epsilon^2
\end{align*}
where the second inequality comes from the definition of $g^{-1}_{\mathcal{X}} (r,\epsilon)$.

Then $\forall x''\in\mathcal{X}$, we have 
\begin{align*}
    \frac{\int_{\overline{g}} (h_1(x)-h_1(x'))^2 dx'}{\int_{\overline{g}} 1 dx'} & \leq C_1^2 \epsilon^2 \\
        &  < \frac{1}{5C_2}\frac{\int_{\mathcal{X}} (h_1(x)-h_1(x''))^2 dx''}{\int_{\mathcal{X}} 1 dx''}\\
    & < \frac{\int_{\mathcal{X}} (h_1(x)-h_1(x''))^2 dx''}{5\int_{\mathcal{X}} 1 dx''}\\
\end{align*}
where the first inequality comes from averaging  $(h_1(x)-h_1(x'))^2$ term on $\overline{g}$, the second inequality comes from the Assumption \ref{ass:gen_variance}.

We have the local generalized variance less than the global generalized variance $\forall x\in\mathcal{X}_i\cap \mathcal{X}_j, x'\in g^{-1}_{\mathcal{X}} (r,\epsilon), x''\in\mathcal{X}$:
\begin{equation}\label{eqn:localvar<globalvar1}
    \frac{\int_{x'\in \overline{g}}(h_1(x)-h_1(x'))^2dx'}{|\overline{g}|} <\frac{1}{5C_2}\frac{\int_{x''\in \mathcal{X}}(h_1(x)-h_1(x''))^2dx''}{|\mathcal{X}|}<\frac{1}{C_2}\frac{\int_{x''\in \mathcal{X}}(h_1(x)-h_1(x''))^2dx''}{|\mathcal{X}|}
\end{equation}
\begin{equation}\label{eqn:localvar<globalvar2}
    \frac{\int_{x'\in \overline{g}}(h_1(x)-h_1(x'))^2dx'}{|\overline{g}|} <\frac{\int_{x''\in \mathcal{X}}(h_1(x)-h_1(x''))^2dx''}{5|\mathcal{X}|}<\frac{\int_{x''\in \mathcal{X}}(h_1(x)-h_1(x''))^2dx''}{|\mathcal{X}|}
\end{equation}


Now we are ready to compare each pair of terms: 

$\bullet$ term $\textcircled{1},\textcircled{2}$:


\begin{align*}
    & \int_{x\in\mathcal{X}}\frac{\int_{x'\in \overline{g}}(h_1(x)-h_1(x'))^2dx'}{\int_{x''\in \mathcal{X}}(h_1(x)-h_1(x''))^2dx''}*\frac{1}{\int_{x''\in \mathcal{X}}(h_1(x)-h_1(x''))^2dx''}dx \\
    \shortintertext{by \Cref{eqn:localvar<globalvar1}}
    & < \int_{x\in\mathcal{X}}\frac{\int_{x'\in \overline{g}}(h_1(x)-h_1(x'))^2dx'}{\int_{x''\in \mathcal{X}}(h_1(x)-h_1(x''))^2dx''}*\frac{1}{C_2}\frac{1}{\int_{x'\in \overline{g}}(h_1(x)-h_1(x'))^2dx'} \frac{|\overline{g}|}{|\mathcal{X}|} dx\\
    & = \int_{x\in\mathcal{X}}\frac{1}{\int_{x''\in \mathcal{X}}(h_1(x)-h_1(x''))^2dx''}\frac{1}{C_2}\frac{|\overline{g}|}{|\mathcal{X}|} dx\\
    & \leq|\mathcal{X}| \frac{1}{|\mathcal{X}|\min_{x\in\mathcal{X}}\int_{x''\in \mathcal{X}}(h_1(x)-h_1(x''))^2dx''}\frac{1}{C_2}\frac{|\overline{g}|}{|\mathcal{X}|} \\
    \shortintertext{by Assumption \ref{ass:gen_variance}}
    &=|\mathcal{X}| \frac{1}{|\mathcal{X}|\max_{x\in\mathcal{X}}\int_{x''\in \mathcal{X}}(h_1(x)-h_1(x''))^2dx''}\frac{|\overline{g}|}{|\mathcal{X}|} \\
    & < \int_{x\in\overline{g}}\frac{1}{\int_{x''\in \mathcal{X}}(h_1(x)-h_1(x''))^2dx''} dx\\
    & = \int_{x\in\overline{g}}\frac{\int_{x'\in \overline{g}}(h_1(x)-h_1(x'))^2dx'}{\int_{x''\in \mathcal{X}}(h_1(x)-h_1(x''))^2dx''}*\frac{1}{\int_{x'\in \overline{g}}(h_1(x)-h_1(x'))^2dx'}dx
\end{align*}
Then we have $\textcircled{1}+\textcircled{2}>0$

$\bullet$ term $\textcircled{3}$:
The following equation is directly from taking the inverse of \Cref{eqn:localvar<globalvar2} when $h_1(x)=s$
\begin{align*}
    & \frac{|\overline{g}|}{\int_{x'\in \overline{g}}(s-h_1(x'))^2dx'}>\frac{|\mathcal{X}|}{\int_{x''\in \mathcal{X}}(s-h_1(x''))^2dx''} \\
\end{align*}

$\bullet$ term $\textcircled{4}$:

\begin{equation}\label{eqn:term4}
\begin{split}
    & \frac{\int_{x''\in \mathcal{X}}(s-h_1(x''))dx''}{\int_{x''\in \mathcal{X}}(s-h_1(x''))^2dx''} / \frac{\int_{x'\in \overline{g}}(s-h_1(x'))dx'}{\int_{x'\in \overline{g}}(s-h_1(x'))^2dx'} \\
    & = \frac{\int_{x''\in \mathcal{X}}(s-h_1(x''))dx''/|\mathcal{X}|}{\int_{x''\in \mathcal{X}}(s-h_1(x''))^2dx''/|\mathcal{X}|} / \frac{\int_{x'\in \overline{g}}(s-h_1(x'))dx'/|\overline{g}|}{\int_{x'\in \overline{g}}(s-h_1(x'))^2dx'/|\overline{g}|} \\
    & = \frac{\int_{x''\in \mathcal{X}}(s-h_1(x''))dx''/|\mathcal{X}|}{\int_{x'\in \overline{g}}(s-h_1(x'))dx'/|\overline{g}|} / \frac{\int_{x''\in \mathcal{X}}(s-h_1(x''))^2dx''/|\mathcal{X}|}{\int_{x'\in \overline{g}}(s-h_1(x'))^2dx'/|\overline{g}|} \\
     \text{by \Cref{eqn:localvar<globalvar2}}\\
    & > \frac{\int_{x''\in \mathcal{X}}(s-h_1(x''))dx''/|\mathcal{X}|}{\int_{x'\in \overline{g}}(s-h_1(x'))dx'/|\overline{g}|} \\
\end{split}
\end{equation}
$\int_{x''\in \mathcal{X}}(s-h_1(x''))dx''/|\mathcal{X}|$ can be seen as the difference between $s$ and the mean of $h_1$ over $\mathcal{X}$.

Now consider two scenarios, 

1. If $\int_{x''\in \mathcal{X}}(s-h_1(x''))dx''/|\mathcal{X}|>C_1\epsilon$, then we have 
\begin{align*}
    \text{\Cref{eqn:term4}} & > \frac{\int_{x''\in \mathcal{X}}(s-h_1(x''))dx''/|\mathcal{X}|}{\int_{x'\in \overline{g}}C_1|r-g(x')|/|\overline{g}|} \\
    & > \frac{\int_{x''\in \mathcal{X}}(s-h_1(x''))/|\mathcal{X}|}{C_1\epsilon} \\
    &>1
\end{align*}
therefore,
\begin{align*}
    4\frac{\int_{x''\in \mathcal{X}}(s-h_1(x''))dx''}{\int_{x''\in \mathcal{X}}(s-h_1(x''))^2dx''}(\frac{\int_{x''\in \mathcal{X}}(s-h_1(x''))dx''}{\int_{x''\in \mathcal{X}}(s-h_1(x''))^2dx''} - \frac{\int_{x'\in \overline{g}}(s-h_1(x'))dx'}{\int_{x'\in \overline{g}}(s-h_1(x'))^2dx'})>0
\end{align*}
With all the summations of each pair of terms in the big bracket being positive, the proof is complete.

2. If $\int_{x''\in \mathcal{X}}(s-h_1(x''))dx''/|\mathcal{X}|\leq C_1\epsilon$, we neglect the positive term ($4\frac{\int_{x''\in \mathcal{X}}(s-h_1(x''))dx''}{\int_{x''\in \mathcal{X}}(s-h_1(x''))^2dx''}\frac{\int_{x''\in \mathcal{X}}(s-h_1(x''))dx''}{\int_{x''\in \mathcal{X}}(s-h_1(x''))^2dx''}$) in $\textcircled{4}$ and compare the negative term ($-4\frac{\int_{x''\in \mathcal{X}}(s-h_1(x''))dx''}{\int_{x''\in \mathcal{X}}(s-h_1(x''))^2dx''}\frac{\int_{x'\in \overline{g}}(s-h_1(x'))dx'}{\int_{x'\in \overline{g}}(s-h_1(x'))^2dx'}$) with the negative terms ($-\frac{|\mathcal{X}|}{\int_{x''\in \mathcal{X}}(s-h_1(x''))^2dx''}$) in $\textcircled{3}$. We have

\begin{equation}\label{eqn:term4_2}
\begin{split}
    & 4\frac{\int_{x''\in \mathcal{X}}(s-h_1(x''))dx''}{\int_{x''\in \mathcal{X}}(s-h_1(x''))^2dx''}\frac{\int_{x'\in \overline{g}}(s-h_1(x'))dx'}{\int_{x'\in \overline{g}}(s-h_1(x'))^2dx'} \\
    & \leq 4\frac{C_1\epsilon|\mathcal{X}|}{\int_{x''\in \mathcal{X}}(s-h_1(x''))^2dx''}\frac{\int_{x'\in \overline{g}}(s-h_1(x'))dx'}{\int_{x'\in \overline{g}}(s-h_1(x'))^2dx'} \\
    & \rightarrow  4\frac{C_1\epsilon|\mathcal{X}|}{\int_{x''\in \mathcal{X}}(s-h_1(x''))^2dx''}\frac{\int_{x'\in \overline{g}}(h_1(x)-h_1(x'))dx'}{\int_{x'\in \overline{g}}(h_1(x)-h_1(x'))^2dx'} \\
    \text{by  \Cref{ass:epsilon}} \\
    & \leq 4\frac{|\mathcal{X}|}{\int_{x''\in \mathcal{X}}(s-h_1(x''))^2dx''}
\end{split}
\end{equation}


Considering $\textcircled{3}+\textcircled{4}$ together, we have 

\begin{align*}
    & \frac{|\overline{g}|} {\int_{x'\in \overline{g}}(s-h_1(x'))^2dx'}-\frac{|\mathcal{X}|}{\int_{x''\in \mathcal{X}}(s-h_1(x''))^2dx''}\\
    &+4\frac{\int_{x''\in \mathcal{X}}(s-h_1(x''))dx''}{\int_{x''\in \mathcal{X}}(s-h_1(x''))^2dx''}\frac{\int_{x''\in \mathcal{X}}(s-h_1(x''))dx''}{\int_{x''\in \mathcal{X}}(s-h_1(x''))^2dx''}-4\frac{\int_{x''\in \mathcal{X}}(s-h_1(x''))dx''}{\int_{x''\in \mathcal{X}}(s-h_1(x''))^2dx''}\frac{\int_{x'\in \overline{g}}(s-h_1(x'))dx'}{\int_{x'\in \overline{g}}(s-h_1(x'))^2dx'}\\
    \shortintertext{neglecting the third term (because it is positive) and by \Cref{eqn:term4_2}}
    & > \frac{|\overline{g}|} {\int_{x'\in \overline{g}}(s-h_1(x'))^2dx'}-\frac{5|\mathcal{X}|}{\int_{x''\in \mathcal{X}}(s-h_1(x''))^2dx''} \\
    \shortintertext{by \Cref{eqn:localvar<globalvar2}}
    & > 0
\end{align*}
Under this scenario, all the summations of each pair of terms in the big bracket are also positive, then we have $\frac{d^2\mathcal{L}}{d\delta^2}>0$, the proof is complete.

\end{proof}

\newpage

\section{Supplementary Experiments}
\subsection{Experiment Details}\label{appendix:experiment details}
\input{tables/experiment_details}
All the dynamical modeling and contrastive learning neural networks mentioned in the paper are fully connected neural networks, with 1 hidden layer of 100 neurons and tanh activations. The autoencoder used for the heat equation is also fully connected, both encoder and decoder use 2 hidden layers (32,16 neurons) and tanh activations.
We conduct all the experiments on a single 2080Ti GPU. The additional experiment details are listed in \Cref{tab:experimenta_details}.

\subsection{Comparison with Other Dynamics Model Baselines}\label{appendix:experiment hnn}

\input{tables/experiment_results_sindy}

We provide a comparison between our model and one of the most popular classical non-DNN based system identification methods called SINDy (Sparse Identification of Nonlinear Dynamics, \citet{brunton2016sindy}) on the performance of dynamics learning. We use the PySINDy package \citep{kaptanoglu2021pysindy} which automatically selects the sparse polynomial basis functions. As expected, SINDy outperforms our DNN based approach on the first two linear examples because they fit in the polynomial basis functions and SINDy is more compact than DNN models. However, SINDy does not perform well on the Kepler example requiring more complex basis functions. Besides, SINDy cannot directly handle the heat equation problem due to the large space dimension.

\input{tables/experiment_results_hnn.tex}
\input{tables/experiment_results_hnn_r_value.tex}
\Cref{tab:experiment_results_HNN} and \Cref{tab:experiment_results_HNN_r_value} show the result comparison between our proposed method and another DNN-based method HNN \citep{greydanus2019hamiltonian} on the two examples. HNN is not applicable to the other two experiments, because they are not Hamiltonian systems.

For dynamical system simulation, the two methods show similar performance, and their conservation and coordinate errors are much smaller than the vanilla neural network. 

For contrastive learning with a single conservation value (e.g. ideal spring mass system), HNN performs slightly better than ConCerNet. For systems with multiple conservation laws, there is no universal metric for conservation because any combination of conservation equations is conserved. ConCerNet empirically learns the Angular momentum function and HNN learns the Hamiltonian value. 

We do not provide comparison experiments with prior works that claim to find the exact solution for multiple conservation laws, as these methods rely on human knowledge of prior information such as fitting function classes. Using the Kepler system as an example, because any combination of angular momentum and energy conservation is also conserved, a natural question will be why the method could find the individual solution instead of a summation of the two. Due to this limitation, it is difficult to evaluate the practical performance of the method because the user/problem-dependent knowledge is unclear.

\subsection{Model Performance with Different Noise Levels}
One might question the ConCerNet performance under large observation noises. We extend the model comparison experiment of \Cref{tab:experiment_results} into various noise settings and show the results in \Cref{tab:experiment_noise}. As the noise level grows, the ConCerNet performance is compromised. However, the baseline NN performance decays much faster than ConCerNet. Especially for potentially unstable systems like the ideal spring mass and Kepler system, both coordinate error and conservation violations could grow exponentially to testing simulation time when training data is corrupted. ConCerNet provides a safety regulation for this issue. Regarding the Heat equation example where the system stabilizes during the simulation, we observe similarly bad performance between the two methods under high noises. This issue might come from the autoencoder learning, as the low representation space cannot interpret the high random noise observation. In the low noise setting, our method can still beat the baseline NN by a large margin.

\input{tables/experiment_noise.tex}

We also test the performance of contrastive conservation learning under varying noise levels and record the $R^2$ coefficient of the linear regression to the exact conservation function in \Cref{tab:noise_r2_value}. As expected, the $R^2$ coefficient decreases with larger noise, but a strong linear association exists for noise standard deviation under 0.1. For the heat equation problem with a noise standard deviation of 1.0, we notice the autoencoder is incapable to capture the major modes in the system, this partially explains the bad performance of both conservation learning and dynamics learning under high noises.

\input{tables/data_r2_noise}




\subsection{Data Efficiency of the Conservation Learning}\label{appendix:data efficiency}
We test the data efficiency of the contrastive conservation learning and record the $R^2$ coefficient of the linear regression to the exact conservation function in \Cref{tab:data_efficiency_r_value}.

\input{tables/data_efficiency_results2.tex}

%% file: tables/experiment_details.tex
\begin{table*}[htb]
\centering
\caption{Experimental details}\label{tab:experimenta_details}
\scalebox{0.8}{
\begin{tabular}{c|c c c c}
        \hline
        & Ideal spring mass  & Chemical kinematics &  Kepler system & Heat equation  \\
        \hline
        Dimension & 2 & 2 & 4 & 101 \\
        Dynamics & \makecell{$\dot{x}[1] = x[2]$\\ $\dot{x}[2] = -x[1]$} & \makecell{$\dot{x}[1] = -\kappa_1 x[1]+\kappa_2 x[2]$\\ $\dot{x}[2] = +\kappa_1 x[1]-\kappa_2 x[2]$} & \makecell{$\dot{x}[1] = x[3]$ \\ $\dot{x}[2] = x[4]$ \\ $r = \sqrt{x[1]^2+x[2]^2}$ \\$\dot{x}[3] = -\frac{x[1]}{r^3}$ \\ $\dot{x}[4] = -\frac{x[2]}{r^3}$ } & $\frac{\partial U}{\partial t}=\frac{\partial^2 U}{\partial y^2}$ \\
        Conservations & $x[1]^2+x[2]^2=C$ & $x[1]+x[2]=C$ &  \makecell{$\frac{x[3]^2+x[4]^2}{2}-\frac{1}{\sqrt{x[1]^2+x[2]^2}}=C_1$ \\ $x[1]x[4]-x[2]x[3]=C_2$} & $\int U dx = C$ \\
        Learning rate & 1e-3 & 1e-3 & 1e-3 & 1e-3\\
        $\# $ of training traj. & 50 & 50 & 50 & 100\\
        Training traj. length (s) & 10 & 10 & 10 & 10\\
        $\# $ of points in training traj. & 100 & 100 & 100 & 100\\
        Training data noise std. & 0.01 & 0.01 & 0.01 & 0.01\\
        $\# $ of eval traj. & 10 & 10 & 10 & 10\\
        Eval. traj. length (s) & 50 & 10 & 5 & 1\\
        \makecell{Constrastive Learning \\ Batch size}& 10 & 10 & 10 & 10 \\
        \makecell{Constrastive Learning \\ Epochs} & 1000 & 1000 & 1000 & 1000\\
        \makecell{Dynamics Learning \\ Batch size}& 100 & 100 & 100 & 100 \\
        \makecell{Dynamics Learning \\ Epochs} & 1000 & 1000 & 1000 & 1000\\
        \hline
\end{tabular}}
\label{tab:experiment_details}
\end{table*}

%% file: tables/experiment_results_sindy.tex
\begin{table*}[htb]
\centering
\caption{Simulation error comparison with classical (non-DNN based) method (SINDy \citep{brunton2016sindy})}
\scalebox{0.8}{
\begin{tabular}{c|c c c c}
        \hline
        \multirow{2}{*}{Task} & \multicolumn{2}{c}{Mean square error} &\multicolumn{2}{c}{Violation of conservation laws}   \\
        &  ConCerNet & SINDy &  ConCerNet & SINDy \\
        \hline
        Ideal spring mass & 9.2e-3 $\pm$ 5.6e-3 & \textbf{1.3e-4$\pm$7.6e-5} & \textbf{1.4e-4 $\pm$ 7.4e-5} & 1.5e-4 $\pm$ 5.6e-5   \\
        Chemical reaction &  5.9e-3$\pm$ 3.7e-3 & \textbf{1.4e-3 $\pm$ 3.6e-4} & 5.6e-3 $\pm$ 5.4e-3   & \textbf{6.6e-7 $\pm$ 3.2e-7}\\ 
        Kepler system &  \textbf{0.140 $\pm$ 0.067} & 1.1 $\pm$ 0.47 &  \textbf{5.2e-3 $\pm$ 4.2e-3}   & 10.2 $\pm$ 6.8\\ 
        \hline
\end{tabular}}
\label{tab:experiment_results_sindy}
\end{table*}

%% file: tables/experiment_results_hnn.tex
\begin{table*}[htb]
\centering
\caption{Simulation error comparison with DNN-based prior work (HNN \citep{greydanus2019hamiltonian})}
\scalebox{0.8}{
\begin{tabular}{c|c c c c c c}
        \hline
        \multirow{2}{*}{Task} & \multicolumn{3}{c}{Mean square error} &\multicolumn{3}{c}{Violation of conservation laws}   \\
        & Baseline NN  & ConCerNet & HNN & Baseline NN  & ConCerNet & HNN \\
        \hline
        Ideal spring mass &0.022 $\pm$ 0.023 & \textbf{9.2e-3 $\pm$ 5.6e-3} & 0.016$\pm$0.024 & 0.012 $\pm$ 0.032 &  1.4e-4 $\pm$ 7.4e-5 &\textbf{2.7e-5 $\pm$ 1.9e-5}\\
        Kepler system & 0.794 $\pm$ 0.476 & \textbf{0.140 $\pm$ 0.067} & 0.174 $\pm$ 0.126& 0.016 $\pm$ 0.011 &  \textbf{5.2e-3 $\pm$ 4.2e-3}   & 0.019 $\pm$ 0.014\\ 
        \hline
\end{tabular}}
\label{tab:experiment_results_HNN}
\end{table*}

%% file: tables/experiment_results_hnn_r_value.tex
\begin{table*}[htb!]
\centering
\caption{Average $R^2$ comparison with prior work (HNN \citep{greydanus2019hamiltonian}) in conservation property learning}
\scalebox{1.0}{
\begin{tabular}{c|c |c|c}
        \hline
        Task & Conservation & ConCerNet & HNN \\
        \hline
        Ideal spring mass system & $x[1]^2+x[2]^2=C$ & $0.997\pm 0.001$& $0.9995\pm 0.0002$\\ 
        Kepler system & $\frac{x[3]^2+x[4]^2}{2}-\frac{1}{\sqrt{x[1]^2+x[2]^2}}=C_1$& $0.858\pm 0.104$& $0.983\pm 0.008$\\
         & $x[1]x[4]-x[2]x[3]=C_2$ & $0.994\pm 0.001$& $0.906\pm 0.032$\\ 
        \hline
\end{tabular}}
\label{tab:experiment_results_HNN_r_value}
\end{table*}

%% file: tables/experiment_noise.tex
\begin{table*}[htb]
\centering
\caption{Simulation error comparison with different noise level}
\scalebox{0.8}{
\begin{tabular}{c|c|c c c c}
        \hline
        \multirow{2}{*}{Task} & \multirow{2}{*}{Noise std.} & \multicolumn{2}{c}{Mean square error} &\multicolumn{2}{c}{Violation of conservation laws}   \\
       & & Baseline NN  & ConCerNet  & Baseline NN  & ConCerNet  \\
        \hline
        \multirow{4}{*}{Ideal spring mass}& 0.0 & 3.7e-3 $\pm$ 2.8e-3 & \textbf{6.1e-4 $\pm$ 5.1e-4} &  0.014 $\pm$ 0.020 &  \textbf{3.5e-5 $\pm$ 4.8e-5} \\
        &0.01 & 0.022 $\pm$ 0.023 & \textbf{9.2e-3 $\pm$ 5.6e-3} &  0.012 $\pm$ 0.032 &  \textbf{1.4e-4 $\pm$ 7.4e-5} \\
        & 0.1 & 0.53 $\pm$ 0.180 & \textbf{0.45 $\pm$ 0.12} &  0.081 $\pm$ 0.033 &  \textbf{1.3e-3 $\pm$ 1.0e-3} \\
       & 1.0 & 21.2 $\pm$ 39.1 & \textbf{0.71 $\pm$ 0.28} &  7.5e4 $\pm$ 1.4e4 &  \textbf{1.45 $\pm$ 2.54} \\
        \hline
        \multirow{4}{*}{Chemical reaction}& 0.0 & 0.01 $\pm$ 0.01 & \textbf{7.3e-3 $\pm$ 9.7e-4} &  0.032 $\pm$ 0.040 &  \textbf{5.5e-3 $\pm$ 4.1e-3} \\
        &0.01 & 9.4e-3 $\pm$ 5.8e-3 & \textbf{5.9e-3 $\pm$ 3.7e-3} &  0.019 $\pm$ 0.026 &  \textbf{5.6e-3 $\pm$ 5.4e-3} \\
        & 0.1 & 0.195 $\pm$ 0.112 & \textbf{0.038 $\pm$ 0.035} &  0.63 $\pm$ 1.02 &  \textbf{0.04 $\pm$ 0.04} \\
       & 1.0 & 2.16 $\pm$ 0.47 & \textbf{0.94 $\pm$ 0.72} &  7.2 $\pm$ 2.6 &  \textbf{1.92 $\pm$ 1.67} \\
       \hline
       \multirow{4}{*}{Kepler system}& 0.0 & 0.13 $\pm$ 0.08 & \textbf{0.10 $\pm$ 0.10} &  5.9e-3 $\pm$ 1.7e-3 &  \textbf{3.7e-4 $\pm$ 3.1e-4} \\
        &0.01 & 0.794 $\pm$ 0.476 & \textbf{0.14 $\pm$ 0.07}& 0.016 $\pm$ 0.011 &  \textbf{5.2e-3 $\pm$ 4.2e-3}  \\
        & 0.1 & 0.89 $\pm$ 0.22 & \textbf{0.73 $\pm$ 0.27} &  0.58 $\pm$ 0.77 &  \textbf{0.50 $\pm$ 0.52} \\
       & 1.0 & 5.27 $\pm$ 3.23 & \textbf{1.21 $\pm$ 0.52} &  144 $\pm$ 90.2 &  \textbf{1.77 $\pm$ 1.51}  \\
       \hline
      \multirow{4}{*}{Heat equation}& 0.0 & 1.2e-4 $\pm$ 2.2e-5 & \textbf{7.2e-5 $\pm$ 1.3e-5} &  0.43 $\pm$ 0.20 &  \textbf{0.027 $\pm$ 0.040} \\
        &0.01 & 7.3e-5 $\pm$ 2.6e-5 & \textbf{2.8e-5 $\pm$ 1.3e-5} & 0.09 $\pm$ 0.07 &  \textbf{0.013 $\pm$ 0.008}  \\
        & 0.1 & 2.5e-4 $\pm$ 3.9e-5 & \textbf{2.4e-4 $\pm$ 9.7e-5} &  \textbf{0.05 $\pm$ 0.03} &  0.11 $\pm$ 0.09 \\
       & 1.0 & \textbf{0.013 $\pm$ 6.2e-4} & 0.013 $\pm$ 2.7e-4 & \textbf{2.95 $\pm$ 0.76} &  3.01 $\pm$ 0.73  \\
       \hline
\end{tabular}}
\label{tab:experiment_noise}
\end{table*}

%% file: tables/data_r2_noise.tex
\begin{table}[htb]
\centering
\caption{Averaged learned invariant fitting error vs noise level}\label{tab:noise_r2_value}
\scalebox{0.9}{
\begin{tabular}{|l|*{5}{c|}}
        \hline
        Task/Conservation & Noise std.& $R^2$&Task/Conservation &Noise std.&$R^2$\\\hline
        \multirow{4}{*}{\makecell{Ideal spring mass\\ Energy conservation}} & 0.0 & 0.998 & \multirow{4}{*}{\makecell{Kepler system\\ Energy conservation}} & 0.0 & 0.828\\\cline{2-3}\cline{5-6}
        &  0.01  & 0.997   &   & 0.01 & 0.858 \\\cline{2-3}\cline{5-6} 
        &  0.1 & 0.986  &  &  0.1 & 0.800\\\cline{2-3}\cline{5-6}
        &  1.0 & 0.061  &  &  1.0 & 0.154\\\hline
        \multirow{4}{*}{\makecell{Chemical reaction\\ mass conservation}} & 0.0 & 1-2e-5 & \multirow{4}{*}{\makecell{Kepler system\\ Angular momentum \\conservation}} & 0.0 & 0.997\\\cline{2-3}\cline{5-6}
        &  0.01  & 1-2e-6   &   & 0.01 & 0.994 \\\cline{2-3}\cline{5-6} 
        &  0.1 & 1-9e-6  &  &  0.1 & 0.812\\\cline{2-3}\cline{5-6}
        &  1.0 & 0.984  &  &  1.0 & 0.175\\\hline
        \multirow{4}{*}{\makecell{Heat equation\\ Energy conservation}} & 0.0 & 1-7e-4 & \multirow{4}{*}{} & \multirow{4}{*}{} & \multirow{4}{*}{}\\\cline{2-3}
        &  0.01  & 1-5e-4   &   & &  \\\cline{2-3} 
        &  0.1 & 0.875  &  &   & \\\cline{2-3}
        &  1.0 & 0.095  &  &   & \\\hline
\end{tabular}}
\end{table}

%% file: tables/data_efficiency_results2.tex
\begin{table}[htb]
\centering
\caption{Averaged learned invariant fitting error vs trajectory number and points per trajectory }\label{tab:data_efficiency_r_value}
\scalebox{0.8}{
\begin{tabular}{|l|*{5}{c|}}
        \hline
        Task/Conservation & \diagbox{points per traj.}{traj.}&5&10&20&40\\\hline
        \multirow{4}{*}{\makecell{Ideal spring mass\\ Energy conservation}} & 5 & 0.547 & 0.828 & 0.971 & 0.989\\\cline{2-6}
        &10  & 0.971   & 0.995  & 0.997 & 0.997 \\\cline{2-6}
        &20 & 0.996  & 0.996 &  0.987 & 0.994\\\cline{2-6}
        &40 & 0.997  & 0.997 &  0.998 & 0.997 \\\cline{2-6}
        &80 & 0.997  & 0.994 &  0.992 & 0.997\\\hline
        \multirow{4}{*}{\makecell{Chemical reaction\\ mass conservation}} & 5 & 1-4.8e-4 & 1-3.3e-4 & 1-1.5e-5 & 1-3.2e-5\\\cline{2-6}
        &10  & 1-1.2e-4   & 1-1.2e-4  & 1-5.6e-5 & 1-6.3e-3\\\cline{2-6}
        &20 & 1-7.1e-4 & 1-2.3e-4 &  1-1.3e-6 & 1-1.3e-6\\\cline{2-6}
        &40 & 1-8.5e-5  & 1-1.1e-4 &  1-2.5e-5 & 1-1.9e-7 \\\cline{2-6}
        &80 & 1-4.1e-5  & 1-7.8e-6 &  1-4.9e-6  & 1-1.1e-6\\\hline
        \multirow{4}{*}{\makecell{Kepler system\\ Energy conservation}} & 5 & 0.007 & 0.300 & 0.170 & 0.477\\\cline{2-6}
        &10  & 0.290   & 0.346  & 0.503 & 0.763\\\cline{2-6}
        &20 & 0.369 & 0.430 &  0.768 & 0.873\\\cline{2-6}
        &40 & 0.393  & 0.748 &  0.815  & 0.863\\\cline{2-6}
        &80 & 0.492  & 0.698 &  0.817  & 0.911\\\hline
        \multirow{4}{*}{\makecell{Kepler system\\ Angular momentum \\conservation}} & 5 & 0.0005 & 0.367 & 0.194  & 0.615\\\cline{2-6}
        &10  & 0.303   & 0.386  & 0.579 &0.902\\\cline{2-6}
        &20 & 0.380 & 0.530 &  0.879 & 0.922\\\cline{2-6}
        &40 & 0.570  & 0.887 &  0.947 & 0.968 \\\cline{2-6}
        &80 & 0.515  & 0.816 &  0.956  & 0.985\\\hline
        \multirow{4}{*}{\makecell{Heat equation\\ Energy conservation}} & 5 & 0.188 & 0.293 & 0.148  & 0.207\\\cline{2-6}
        &10  & 0.301   & 0.217  & 0.132 &0.318\\\cline{2-6}
        &20 & 0.280 & 0.312 &  0.409 & 0.438\\\cline{2-6}
        &40 & 0.723 & 0.273 &  0.550 & 0.674 \\\cline{2-6}
        &80 & 0.994  & 0.998 &  0.999  & 0.999\\\hline
\end{tabular}}
\end{table}